\documentclass{article}
\usepackage{authblk}
\usepackage{subcaption}
\usepackage[english]{babel}

\usepackage[letterpaper,top=2cm,bottom=2cm,left=3cm,right=3cm,marginparwidth=1.75cm]{geometry}

\usepackage{amsmath}
\usepackage{graphicx}
\usepackage[colorlinks=true, allcolors=blue]{hyperref}

\begin{document}
\title{Robustness in sparse artificial neural networks trained with adaptive topology}
\author[1]{Bendegúz Sulyok}
\author[1,2]{Gergely Palla}
\author[3]{Filippo Radicchi}
\author[3]{Santo Fortunato}
\affil[1]{Dept. of Biological Physics, Eötvös Loránd University, Budapest, Hungary}
\affil[2]{Semmelweis University, Faculty of Health and Public Administration, Health Services Management Training Centre, Budapest, Hungary}
\affil[3]{Center for Complex Networks and Systems Research
Luddy School of Informatics, Computing, and Engineering
Indiana University Bloomington, USA}

\maketitle

\begin{abstract}
%In this paper, w
We investigate the robustness of sparse artificial neural networks trained with adaptive topology. 
%Our study focuses 
We focus
on a simple yet effective architecture consisting of three sparse layers with 99\% sparsity followed by a dense layer, applied to image classification tasks such as MNIST and Fashion MNIST. By updating the topology of the sparse layers between each epoch, we achieve competitive accuracy despite the significantly reduced number of weights. Our primary contribution is a detailed analysis of the robustness of these networks, exploring their performance under various perturbations including random link removal, adversarial attack, and link weight shuffling. Through extensive experiments, we demonstrate that adaptive topology not only enhances efficiency but also maintains robustness. This work highlights the potential of adaptive sparse networks as a promising direction for developing efficient and reliable deep learning models.
\end{abstract}

\section{Introduction}

Deep Neural Networks (DNNs) have become the cornerstone of modern artificial intelligence, achieving state-of-the-art performance in diverse domains such as computer vision, natural language processing, and scientific computing \cite{DBLP_KrizhevskySH12,NIPS_Vaswani}. This success has been fueled by a trend of ever-increasing model size and complexity. Modern models often contain billions of parameters, leading to substantial computational and  memory burden as well as 
%and 
significant
energy cost for both training and inference \cite{Strubell_2019_energy}. This over-parameterization not only poses significant challenges for deployment on resource-constrained devices like mobile phones and IoT sensors, but also raises environmental concerns regarding the carbon footprint of AI.

In parallel, the human brain provides an exceptionally efficient and powerful computational system for which neurological studies have long established that its connectivity is remarkably sparse \cite{Ramon_y_Cajal,Braitenberg_book}. A typical neuron connects to only a %small 
vanishing
fraction of other neurons, forming a complex yet highly efficient network structure. This biological sparsity is believed to be a key principle enabling the brain to learn and perform complex tasks with unparalleled energy efficiency \cite{Braitenberg_book,Hassibi1993OptimalBS,Olhausen_and_Field_1997}. This contrast between dense, computationally heavy artificial networks and the sparse, efficient biological brain drew considerable scientific interest towards sparsity in deep learning \cite{LeCun_opt_brain_damage,Han_efficient_neural_net,Frankle_Lottery_Ticket_Hypothesis,Li_AdaPrune,Sun_pruningapproach,Frantar_sparse_languagemodels,Carlo_post-training,Tegmark_brain_inspired}. 

However, it is worth noting that while artificial Sparse Neural Networks (SNNs) are significantly less connected than traditional dense models, 
they still retain a finite fraction of all the possible edges, meaning that they are not ``sparse'' in the jargon of network science~\cite{newman2018networks}. On the other hand, real neural networks are sparse, as they display only a vanishing fraction of all possible connections. 
%they typically remain orders of magnitude denser than biological networks. 
Despite this distinction, SNNs have emerged as a promising solution to mitigate the burdens of over-parameterisation and are attractive for deployment on resource-constrained devices. By ensuring that a significant portion of the network's weights are zero, SNNs can theoretically achieve dramatic reductions in storage requirements and computational complexity.
%, as multiplications with zero can be skipped entirely

%In general, their potential to reduce computational costs and memory requirements, makes Sparse Neural Networks (SNNs) attractive for deployment on resource-constrained devices. In addition, SNNs, where a significant portion of the network's weights are zero, have emerged as a promising solution to mitigate the burdens of over-parameterization. By eliminating redundant parameters, SNNs can theoretically achieve dramatic reductions in storage requirements and computational complexity, as multiplications with zero can be skipped entirely. 

Seminal work in this area demonstrated that the sparsification of pre-trained dense models via network pruning can be achieved without a significant loss in the accuracy \cite{LeCun_opt_brain_damage,Han_efficient_neural_net}. More recent discoveries, such as the Lottery Ticket Hypothesis, suggest that sparse sub-networks with exceptional potential performance exist within dense models from the very beginning of the training \cite{Frankle_Lottery_Ticket_Hypothesis}. Accordingly, in the pruning approach to SNN 
%we start 
one starts
from a fully connected neural network and gradually removes links while aiming to achieve an optimal balance between final performance and post-training inference speed~\cite{Li_AdaPrune,Sun_pruningapproach,Frantar_sparse_languagemodels,Carlo_post-training}.

An equally important alternative to network pruning is the idea of sparse training, where a sparsely connected network is initiated, ensuring an inherent sparsity throughout the training \cite{Mocanu_2017_dyn_training_Nat_Comms,Lee2018snip,EvciGMCE20,Yuan_2021_MEST,Yuan_2022_layerfreezingdata,Zhang_2022_OptG,Zhang_2023_bidirectionalmasks}. A prominent approach in this direction is dynamic sparse training \cite{Mocanu_2017_dyn_training_Nat_Comms,EvciGMCE20,Yuan_2021_MEST}, which provides an efficient framework, where the network structure is also dynamically evolving during training. Epitopological learning is a variation of dynamic sparse training that is inspired by the plasticity of the brain and the way organic neural networks change their connectivity during learning, forming epitopological engrams (memory traces)~\cite{Carlo_2015_linkpred,Carlo_2017_pioneering,Narula_2017_epitopological,Carlo_2018_brain_inspired}. A relatively simple implementation of epitopological learning consists in applying link prediction during training, where 
%we evaluate 
one evaluates
the likelihood of the existence of missing links based on the current network structure and takes this likelihood into account before the next modification to the network topology. In a slightly more general framework, epitopological learning can be implemented with the help of network automatons that provide connectivity predictions based on the input knowledge and the topological network organization. 

Recently, a sparse network trained in the framework of Epitopological Sparse Meta-deep Learning (ESML) was shown to outperform fully connected networks across multiple architectures and datasets while retaining only 1\% of the connections~\cite{CHCL3_main_paper}. The basis of the framework was provided by the Cannistraci-Hebb (CH) automata learning theory, and the training procedure relied heavily on the CH3-CL network automata rules~\cite{CHCL3_base}, offering an effective solution for general-purposed link prediction in bipartite networks. 

In the present work, we examine this highly effective sparse neural network from the perspective of robustness. In the light of practical applications, robustness in neural networks is a critical concern, as models must perform reliably under various real-world conditions, including the presence of noise, adversarial perturbations, and shifts in data distribution. While sparsity can reduce overfitting and improve generalization, its impact on robustness is less understood, especially in the context of adaptive topologies. Here, we study the resilience of sparse neural networks trained in the ESML approach against various link removal procedures.

%\color{red}
%Sparse neural networks have gained significant attention due to their potential to reduce computational costs and memory requirements, making them attractive for deployment on resource-constrained devices. Traditional approaches to sparsity involve pruning dense networks after training, but recent research has explored training sparse networks from scratch, often with adaptive topologies that evolve during training. This adaptive approach can lead to more efficient and potentially more robust models.

%In this work, we delve into the robustness of such adaptive sparse networks. Specifically, we examine an architecture with three highly sparse layers (99\% sparsity) followed by a dense layer, trained on standard image classification benchmarks, MNIST and CIFAR10. Our training methodology includes updating the sparse topology between epochs, allowing the network to adapt its structure dynamically.

%Robustness in neural networks is a critical concern, as models must perform reliably under various real-world conditions, including the presence of noise, adversarial perturbations, and shifts in data distribution. While sparsity can reduce overfitting and improve generalization, its impact on robustness is less understood, especially in the context of adaptive topologies.
%\color{black}

\section{Results}

\subsection{Sparse architecture and training}

\begin{figure}
    \centering
    \includegraphics[width=0.9\linewidth]{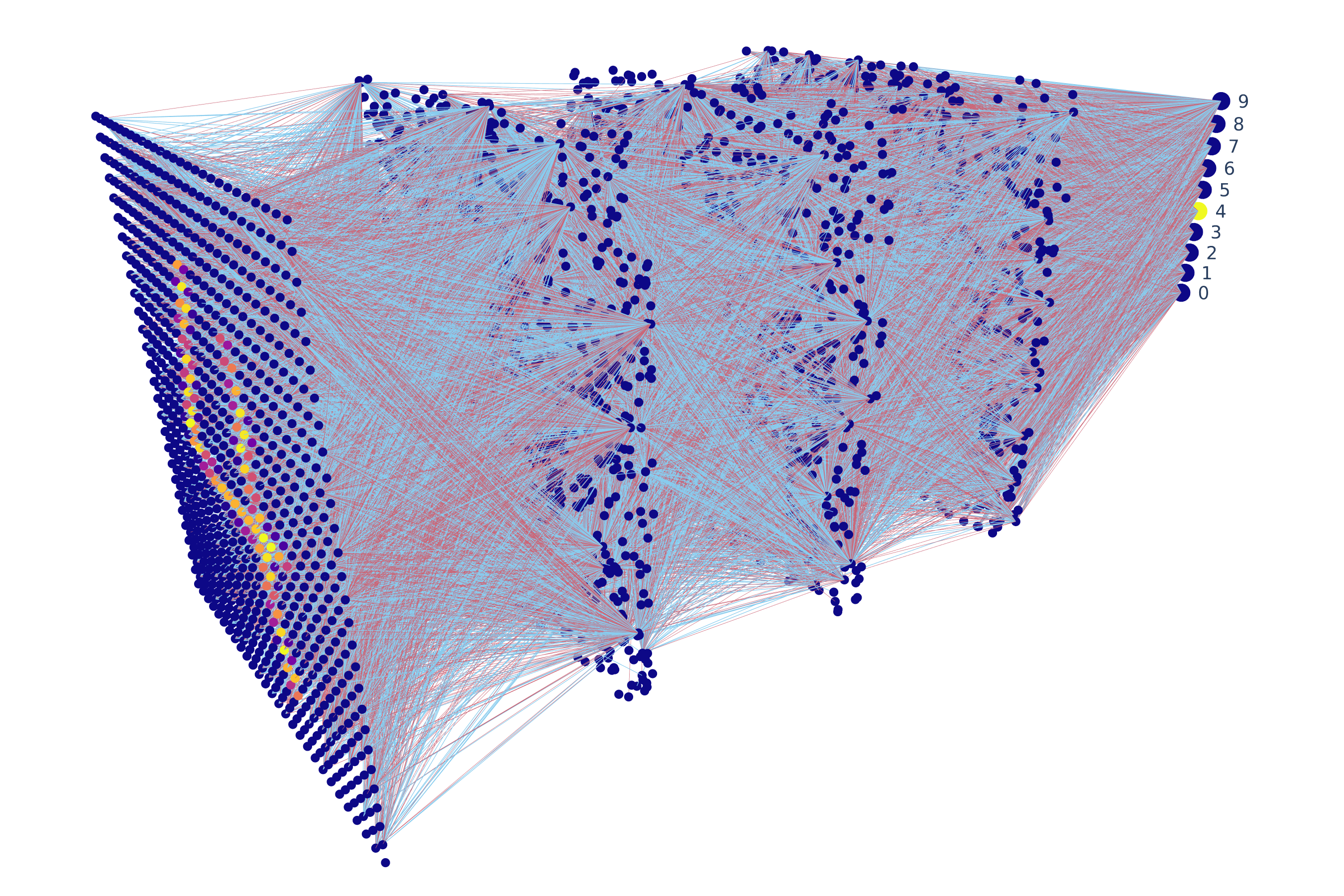}
    \caption{\textbf{Illustration of the network architecture.} The input layer (shown on the left) contains  28 x 28 pixels, followed by 3 sparsely connected neuron layers with 1000 neurons each. Blue coloured connections have a positive link weight, whereas red colour indicates a negative weight. The last layer, providing the readout is densely connected to the third sparse layer.}
    \label{fig:spare_network_structure_plot}
\end{figure}

%We evaluated the proposed training framework using three different strategies for link regrowth during the topology update step. For each strategy, we trained the network across four independent runs. Test accuracy was recorded after every epoch.
The structure of the network architecture we used followed the setup proposed in Ref.~\cite{CHCL3_main_paper}, as illustrated in Fig.~\ref{fig:spare_network_structure_plot}, showing the final stage of the training in one of our experiments with the MNIST dataset. The input layer consists of 784 pixels, which is followed by 3 sparsely connected neuron layers with 1000 neurons each. The network also contains an additional layer for readout, containing 10 neurons that are densely connected to the third layer.

The training of the network starts already with a sparse configuration, where only 1\% of the possible connections are present between the input and the first 3 neuron layers, placed uniformly at random with a weight drawn from a normal distribution having a mean $\mu=0.0$ and a standard deviation $\sigma=\sqrt{\frac{2}{f_{in}}}$, where $f_{in}$ is the size of the previous neuron layer. During the training in each epoch, first the weights are adjusted based on backpropagation. This is followed by rewiring of the connections, where first a small fraction of the links are deleted, and then an equal amount of new links are introduced to keep the overall number of connections in the network constant. 

For the link regrowth procedure we applied two distinct strategies: Random Link Regrowth (RLR) and the CH3L3 heuristic~\cite{CHCL3_base,CHCL3_main_paper}. The RLR method, as its name suggests, assigns the new links simply at random. In contrast, the CH3L3 method is based on link prediction, where the new links are placed according to the highest likelihoods for unobserved connections in the current network topology. The two possible link regrowth strategies were not ``mixed,'' i.e., for any given experiment, one of the two possibilities was used exclusively throughout the entire training.  
\begin{figure}[hbt]
    \centering
    \includegraphics[width=\textwidth]{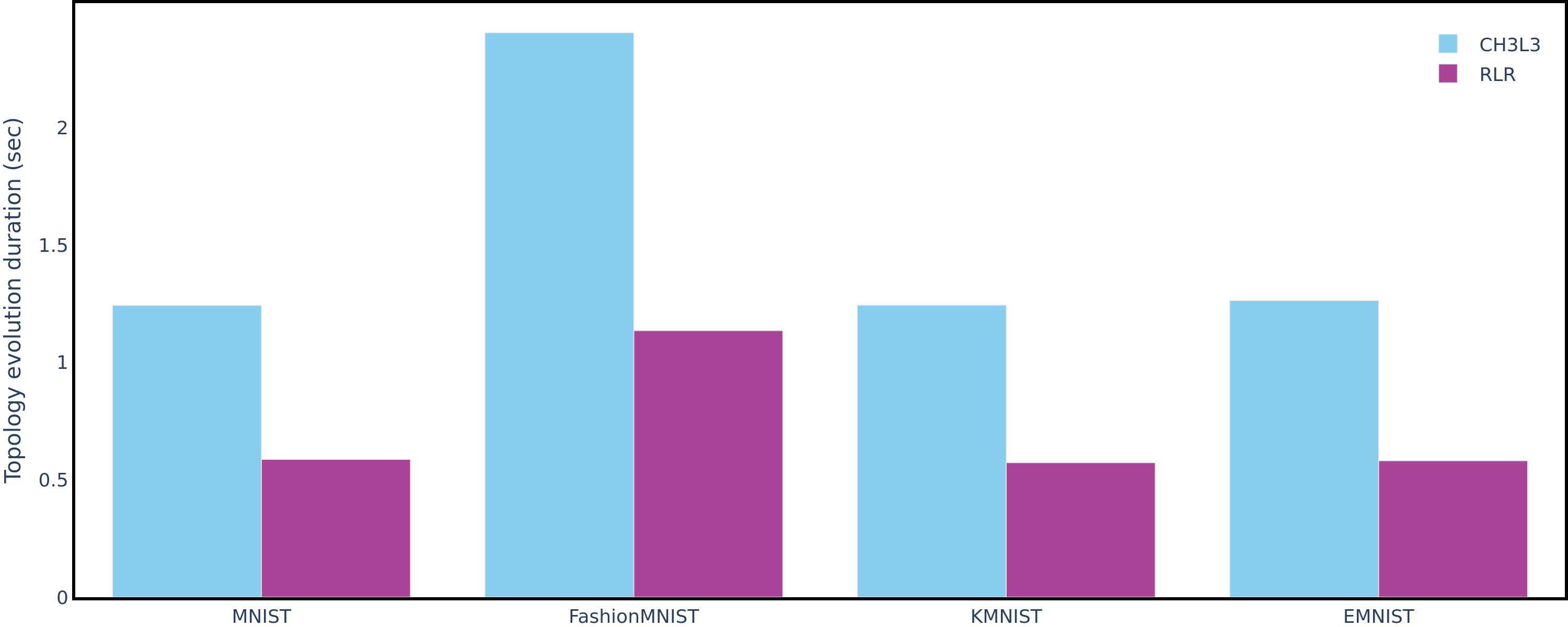}
    \caption{\textbf{Duration of the topology update.} We show the bar chart of the average duration in seconds for a single topology update over 4 training instances and 50 epochs. The results for the CH3L3 method are shown in light blue, whereas for the Random Link Regrowth method in purple, where the dataset is marked under the corresponding bars.}
    \label{fig:topology_evolution_time}
\end{figure}
In Fig.~\ref{fig:topology_evolution_time}, we show the duration of the topology update for the two different link regrowth strategies. As expected, due to its simplicity, the %Random Link Regrowth 
RLR
method roughly halves the average topology update time compared to the more complex CH3L3 method for all studied datasets. For a more detailed description of the training procedures, see Methods.

\subsubsection{Accuracy}
\begin{figure}[htbp]
    \centering
    \begin{subfigure}[t]{0.49\textwidth}
        \caption{MNIST\label{fig:subfigA}}
        \includegraphics[width=\textwidth]{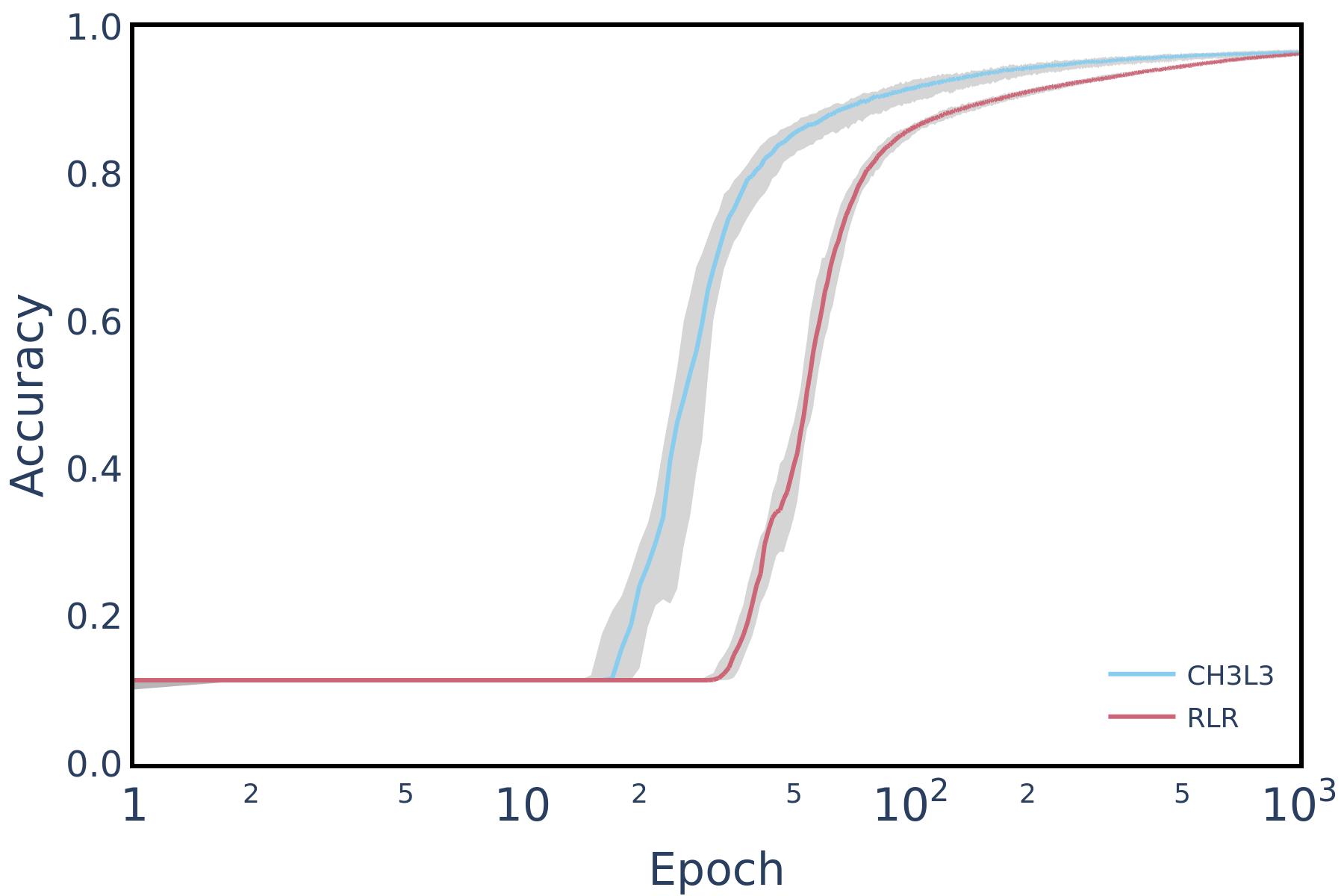}
    \end{subfigure}
    \begin{subfigure}[t]{0.49\textwidth}
        \caption{FashionMNIST\label{fig:subfigB}}
        \includegraphics[width=\textwidth]{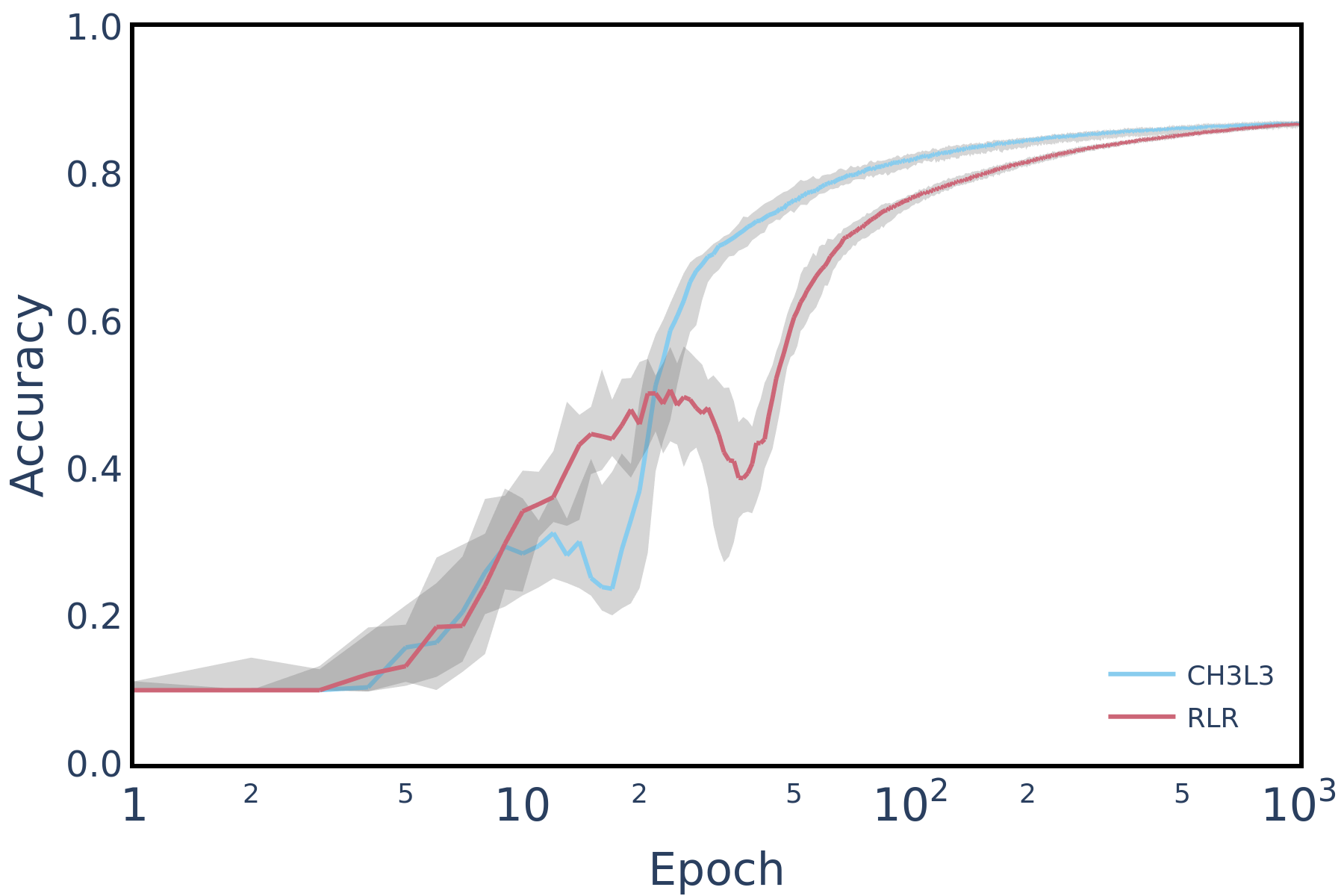}
    \end{subfigure}
    
    \begin{subfigure}[t]{0.49\textwidth}
        \caption{KMNIST\label{fig:subfigC}}
        \includegraphics[width=\textwidth]{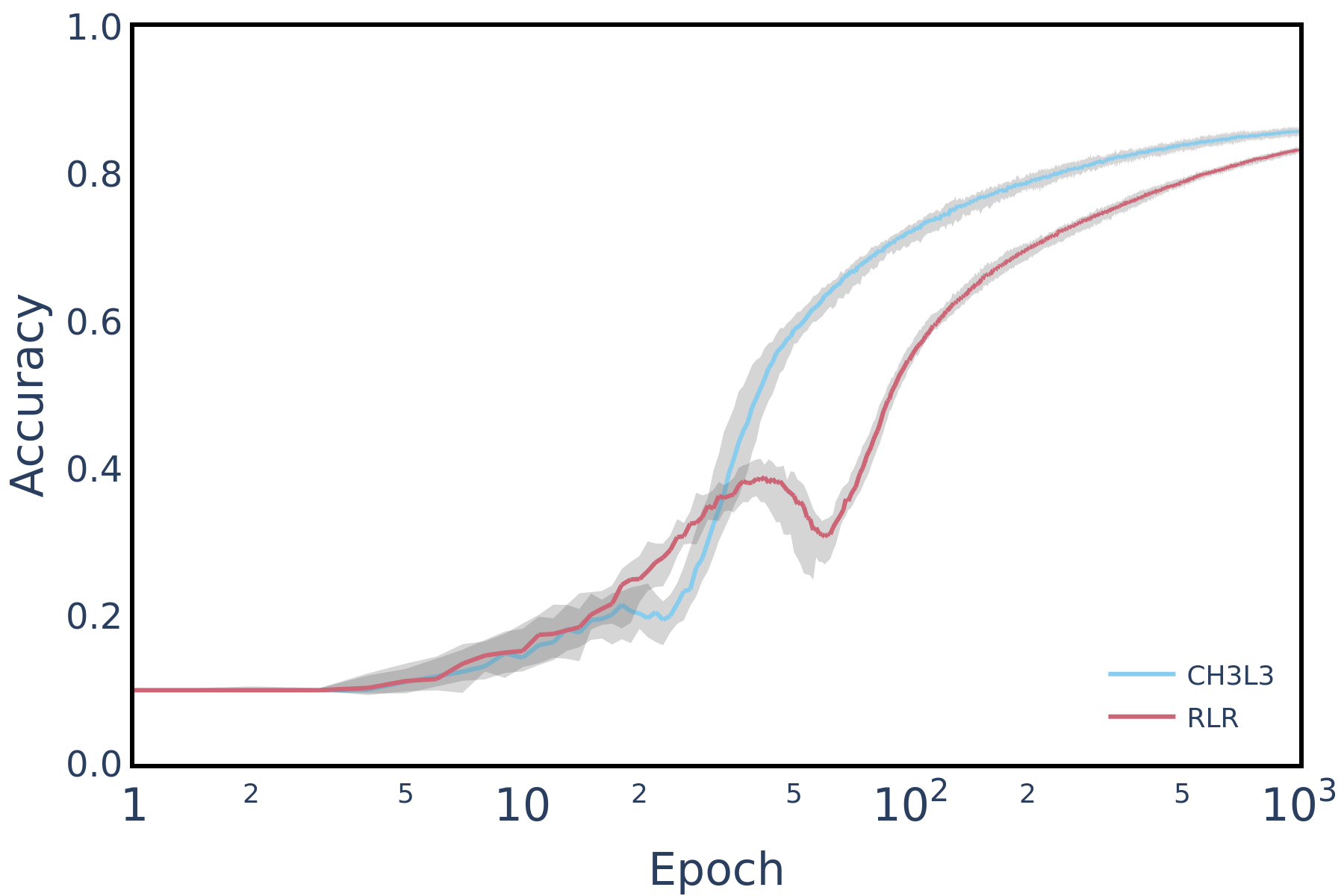}
    \end{subfigure}
    \begin{subfigure}[t]{0.49\textwidth}
        \caption{EMNIST (Letters)\label{fig:subfigD}}
        \includegraphics[width=\textwidth]{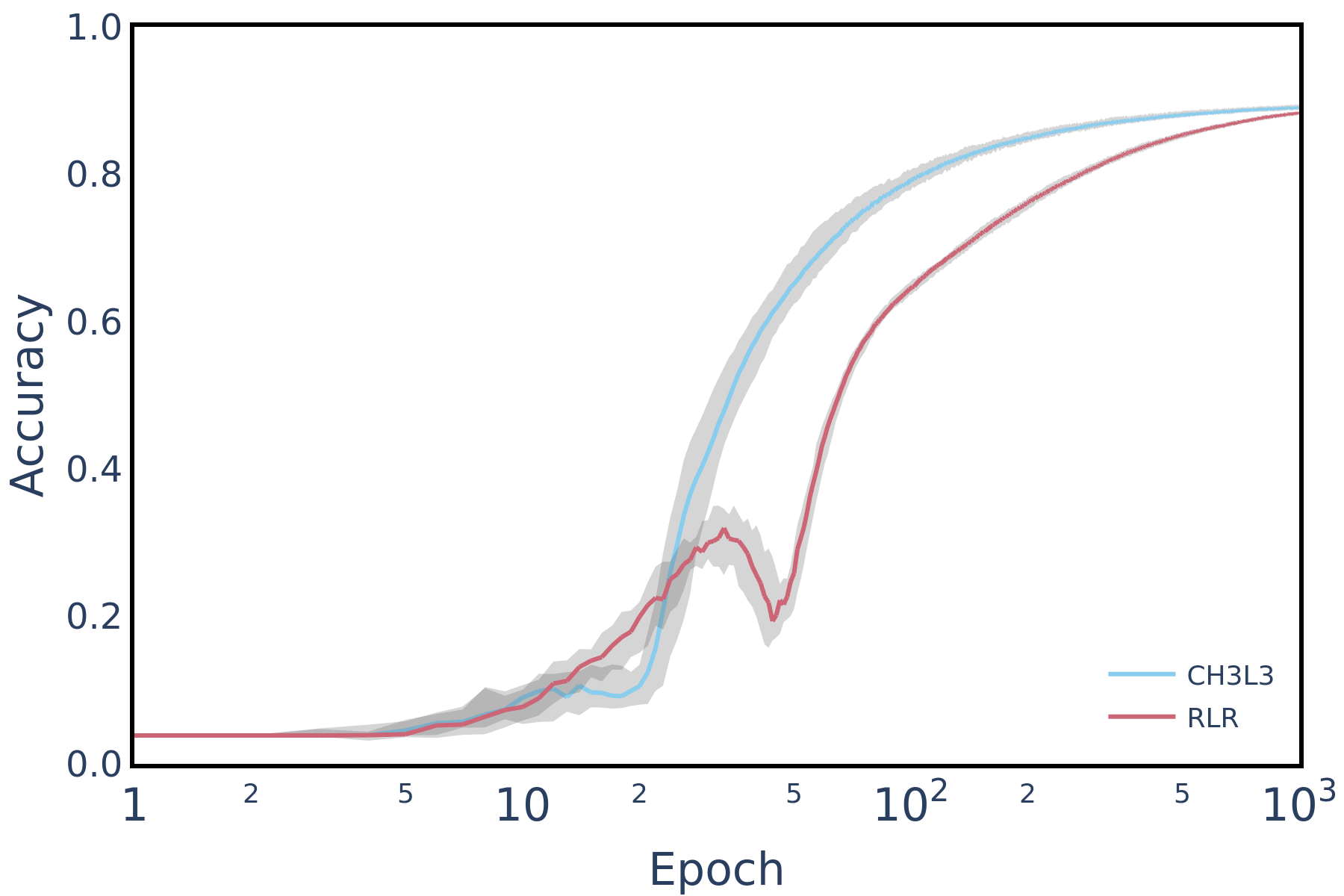}
    \end{subfigure}
    \caption{\textbf{Accuracy during training.} We show the accuracy (
    %percentage 
    fraction
    of correctly classified inputs, where the shaded area indicates the standard deviation) as a function of the number of epochs (on logarithmic scale). The results for the \textit{CH3L3} method are shown in blue, for the \textit{Random Link Regrowth} method in red. Results are displayed for the following datasets: a) MNIST, b) FashionMNIST, c) KMNIST, and d) EMNIST Letters. Each setup was run 10 times with a different random seed and each training instance run for 1000 epochs with a topology update between consecutive epochs.}
    \label{fig:training_curves}
\end{figure}

Figure \ref{fig:training_curves} shows the evolution of test accuracy as a function of the training epochs for the MNIST  (Fig.\ref{fig:training_curves}a), Fashion MNIST (Fig.\ref{fig:training_curves}b), KMNIST (Fig.\ref{fig:training_curves}c), and EMNIST letters datasets (Fig.\ref{fig:training_curves}d), where each curve represents the median over 10 independent training runs started from a different random seed. The shaded region around the curves falls between the  40$^{\rm th}$ and 60$^{\rm th}$ percentiles and indicates that a noticeable variability can be introduced by random initialization in the middle range of the training procedure for some systems. 

\captionsetup[subfigure]{justification=raggedright,singlelinecheck=off}

For the studied datasets, the accuracy of the networks trained according to the CH3L3 method approaches its limiting value considerably faster compared to the accuracy of the networks trained with random link regrowth. However, the accuracy at the end of the training procedure is the same for both approaches within a short margin of uncertainty for all datasets except the KMNIST dataset (Fig.\ref{fig:training_curves}c), where the maximal accuracy of the network trained with the CH3L3 method is slightly higher. Interestingly, in the case of the Fashion MNIST dataset (Fig.\ref{fig:training_curves}b), after hitting a local maximum, the accuracy curve briefly decreases until reaching a local minimum, and then monotonically improves with each further training epoch for both training methods. 

Put together, these results demonstrate that both training methods benefit from the alternating approach between weight update and link rewiring, and the applied sparse dynamic training framework is effective. In the meantime, the specific design of the update rule plays an important role in determining the speed of convergence.

\subsection{Robustness analysis}

We assessed the robustness of the trained sparse networks through a systematic post-training perturbation analysis. After the completion of the training phase, we exposed the networks to two distinct types of disruption: iterative structural pruning (removing connections) and stochastic weight perturbation (randomly altering trained weights). The network performance was re-evaluated on the test set at each perturbation step, critically, without any subsequent fine-tuning or retraining.

%To assess the robustness of the trained sparse networks, we performed a post-training perturbation analysis of different kinds. After completing the training, we applied additional procedures that either subsequently removed fractions of the connections, or changed the trained weights in a random fashion. At each stage of the perturbation process, we re-evaluated the networks on the test set without retraining.
%All perturbations are applied layer-wise.

The detailed list of the applied different types of perturbations is:
%can be given as follows:
\begin{itemize}
    \item \textit{Random Pruning} - links are removed from the network layer-by-layer uniformly at random. %The value is averaged over several measurements.
    \item \textit{Weight Order Pruning} - links are removed by decreasing order of magnitude (i.e., highest-magnitude connections are removed first).
    \item \textit{Reverse Weight Order Pruning} - links are removed in increasing order based on their weight magnitude.
    \item \textit{Weight Shufling} - weight values are shuffled within fixed-size bins applied layer-wise. The size of the perturbation is controlled by a value ranging from $0$ to $1$, which represents the ratio of the bin size by the total range of weight values within that layer. As this value increases, the bins become larger, allowing more disparate weight values to be shuffled together, thereby increasing the strength of the perturbation
    %weight values are layer wise shuffled in fixed sized bins. The perturbation value ranges from $0$ to $1$ and this corresponds to the ratio between the bin size and the range of the weights in the given layer. With this value increasing, the bins become larger and more distant weight values are being shuffled, thus making the perturbation stronger.
    \item \textit{Weight Modification} - random noise, sampled from a normal distribution centered on $0$, is added to the weight values. The perturbation value $m_p$ controls the standard deviation of the noise as $\sigma_p = \Bar{w} \cdot m_p$, where $\Bar{w}$ is the average weight magnitude for the given layer. %and $m_p$ is the perturbation value that is also being displayed on Figure \ref{fig:robustness_analysis_noising}.
\end{itemize}

\captionsetup[subfigure]{justification=raggedright,singlelinecheck=off}
\begin{figure}[htbp]
    \centering
    \begin{subfigure}[t]{0.49\textwidth}
        \caption{MNIST\label{fig:subfigA2}}
        \includegraphics[width=\textwidth]{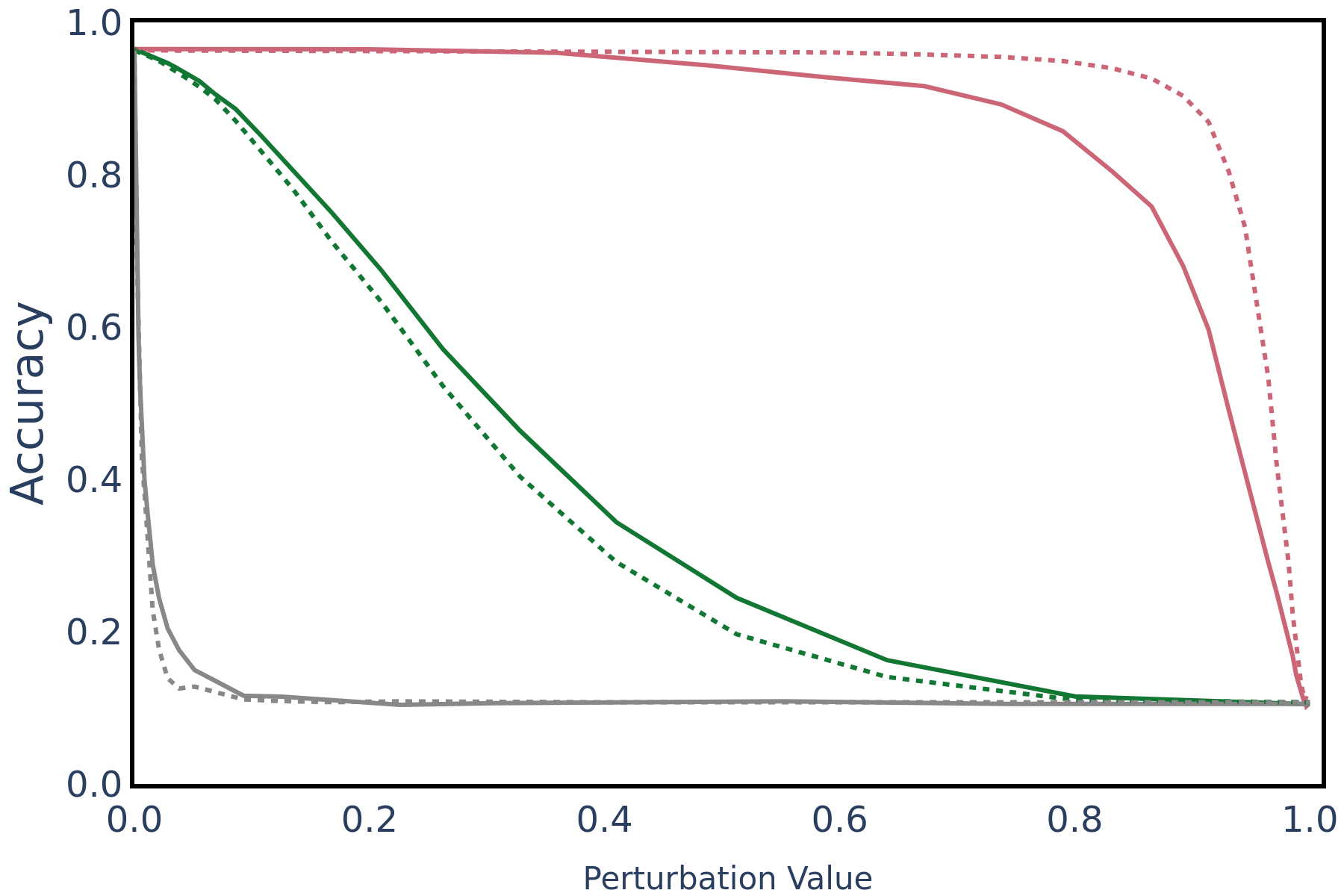}
    \end{subfigure}
    \begin{subfigure}[t]{0.49\textwidth}
        \caption{FashionMNIST\label{fig:subfigB2}}
        \includegraphics[width=\textwidth]{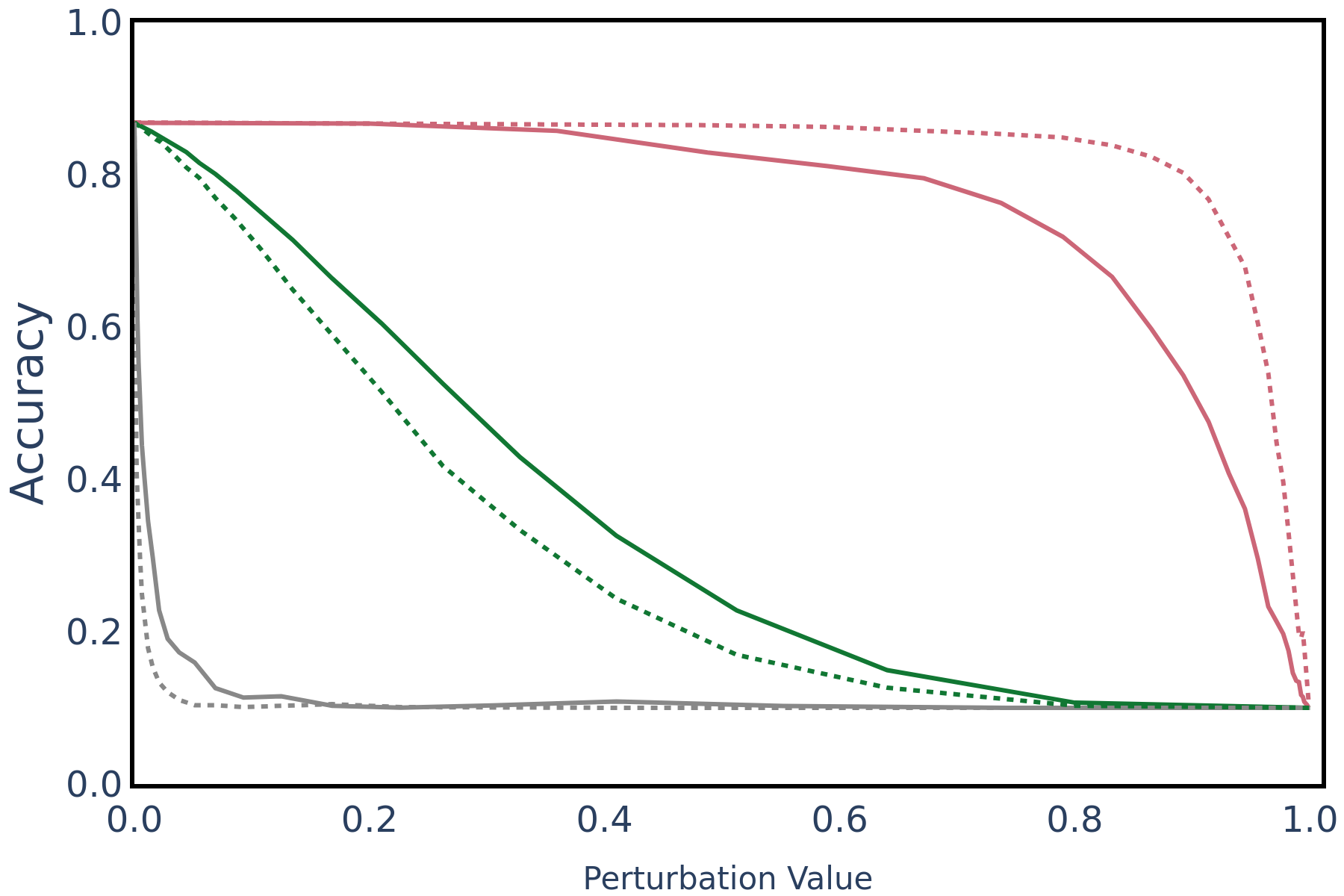}
    \end{subfigure}
    
    \begin{subfigure}[t]{0.49\textwidth}
        \caption{KMNIST\label{fig:subfigC2}}
        \includegraphics[width=\textwidth]{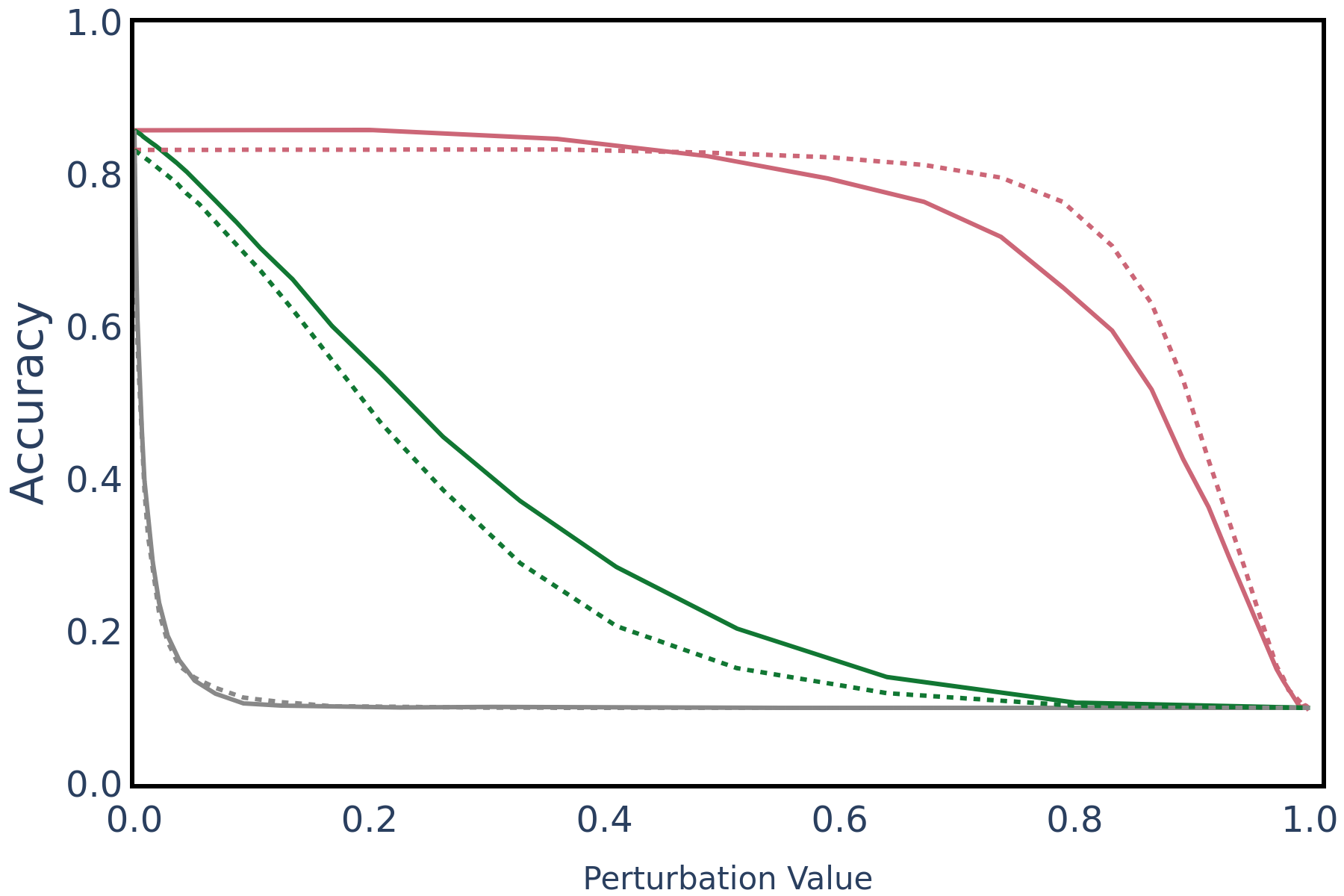}
    \end{subfigure}
    \begin{subfigure}[t]{0.49\textwidth}
        \caption{EMNIST (Letters)\label{fig:subfigD2}}
        \includegraphics[width=\textwidth]{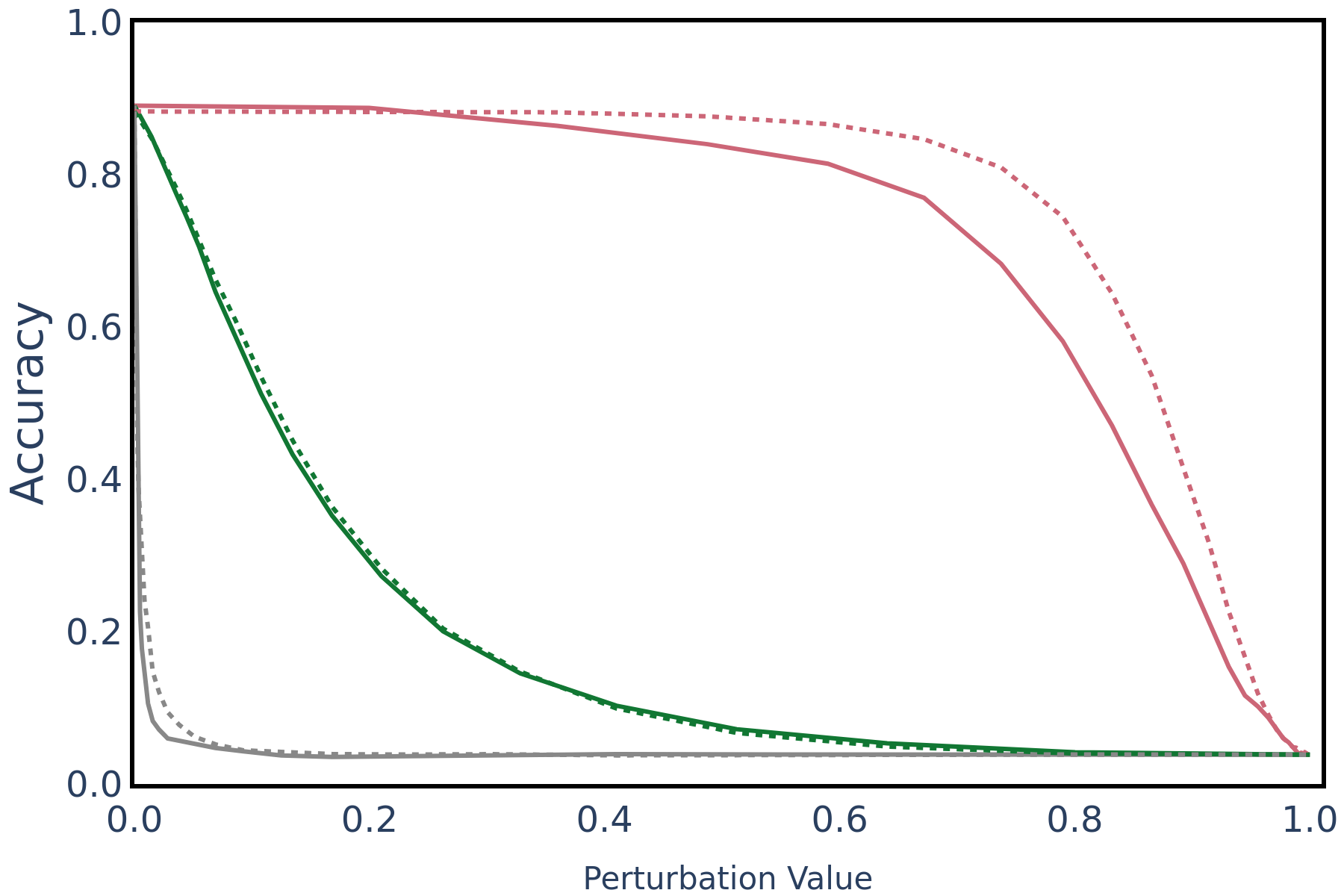}
    \end{subfigure}
    
    \begin{subfigure}[t]{\textwidth}
        \includegraphics[width=\textwidth]{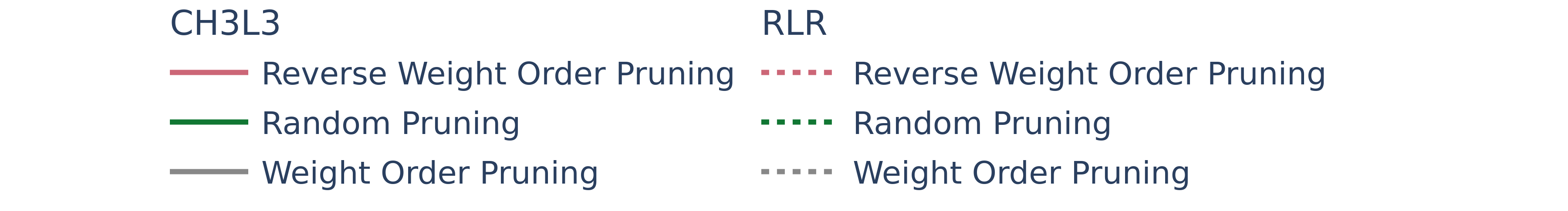}
    \end{subfigure}
    \caption{\textbf{Robustness of trained sparse networks against link removal.} Average accuracy over 32 networks as a function of the fraction of removed links (perturbation value) for systems trained according to the CHCL3 method (continous lines) and the RLR method (dashed lines) for a) the MNIST dataset, b) the Fashion MNIST dataset, c) the KMNIST dataset, and d) the EMNIST letters dataset. Results for Reverse Weight Order Pruning are shown in red, for Random Pruning in green and for Weight Order Pruning in gray.}
    \label{fig:robustness_analysis_pruning}
\end{figure}

In Fig.\ref{fig:robustness_analysis_pruning} we show the accuracy as a function of the fraction of removed links for Random Pruning (green), Weight Order Pruning (gray) and Reverse Weight Order Pruning (red). As expected, for all tested datasets, the Weight Order Pruning decreases the performance of the networks in the most rapid fashion, with an almost vertical drop in the accuracy close to the origin, reaching the minimum value already at 1 to 5\% of removed links. The studied sparse neural networks show more resilience against Random Pruning, where we can observe a gradual decrease in accuracy with roughly a constant (and finite) slope in the early stages of the link removal process, with steepness of the curves slightly reduced in the later stages, eventually decreasing to the minimum at around 80\% removed links. Finally, the studied sparse dynamic neural networks show high resilience against Reverse Weigh Order pruning, where accuracy may suffer only a minor decrease even at 80\% of the links removed, which then is followed by a steep decline for larger fractions of removed links.

When comparing the robustness of the networks from the point of view of the two different training strategies, we can observe that networks trained with the RLR method (dashed lines) show higher resilience against Reverse Weight Order pruning for all datasets in Fig.~\ref{fig:robustness_analysis_pruning}. Resilience against Random Pruning yielded mixed results. Networks trained with the CH3L3 regrowth method showed slightly higher resilience across the MNIST, Fashion MNIST, and KMNIST datasets  (Fig.\ref{fig:robustness_analysis_pruning}a-c). However, for the EMNIST letters dataset, the resilience curves for the two training strategies were nearly indistinguishable (Fig.\ref{fig:robustness_analysis_pruning}d). Finally, performance under Weight Order Pruning was dependent on the dataset. Networks trained using CH3L3 appeared slightly more resilient --though by a very small margin-- for the MNIST and Fashion MNIST datasets (Fig.\ref{fig:robustness_analysis_pruning}a-b). Conversely, the RLR method resulted in slightly higher resilience for the KMNIST and EMNIST letters datasets (Fig.\ref{fig:robustness_analysis_pruning}c-d).

\begin{figure}[htbp]
    \centering
    \begin{subfigure}[t]{0.49\textwidth}
        \caption{MNIST\label{fig:subfigA3}}
        \includegraphics[width=\textwidth]{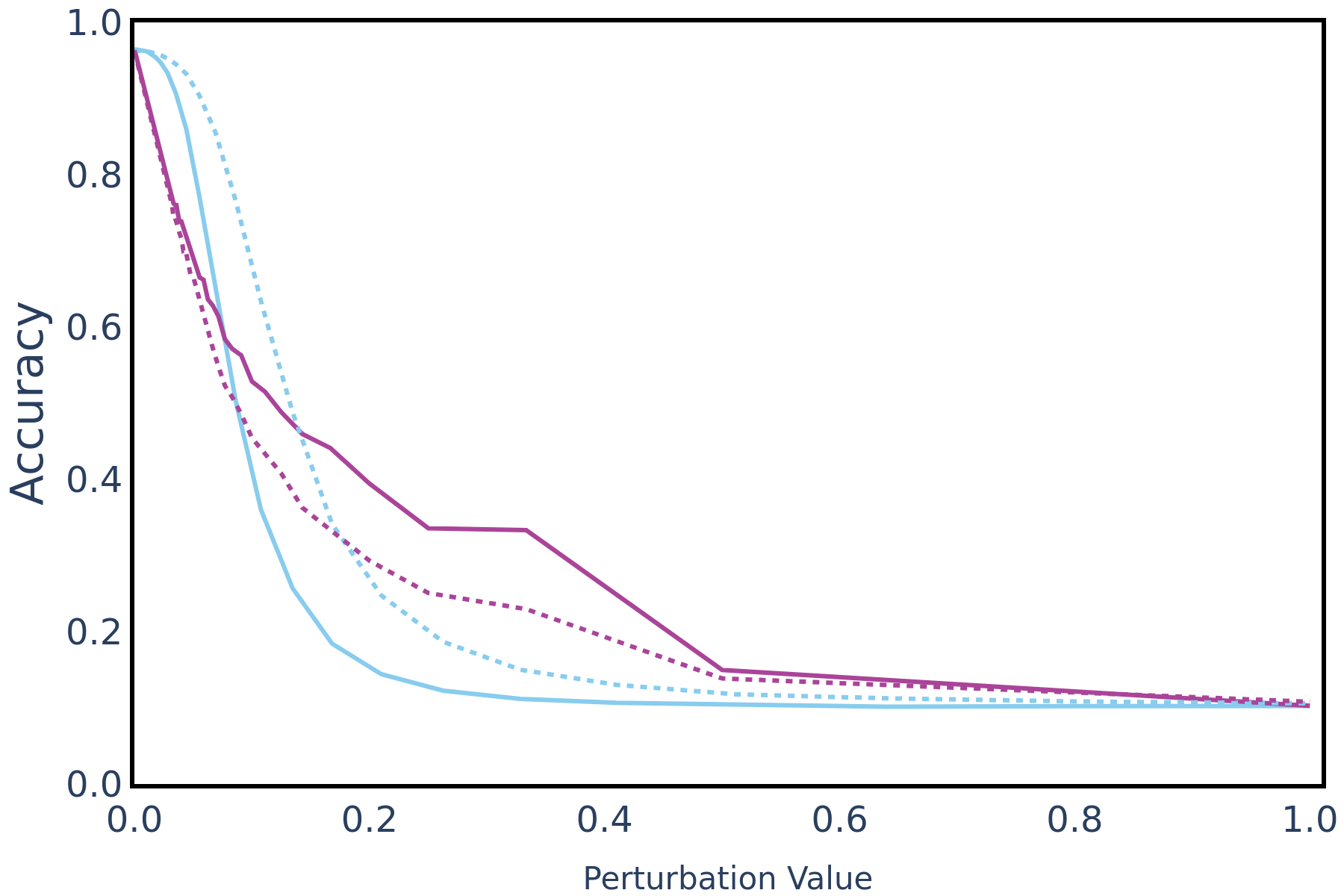}
    \end{subfigure}
    \begin{subfigure}[t]{0.49\textwidth}
        \caption{FashionMNIST\label{fig:subfigB3}}
        \includegraphics[width=\textwidth]{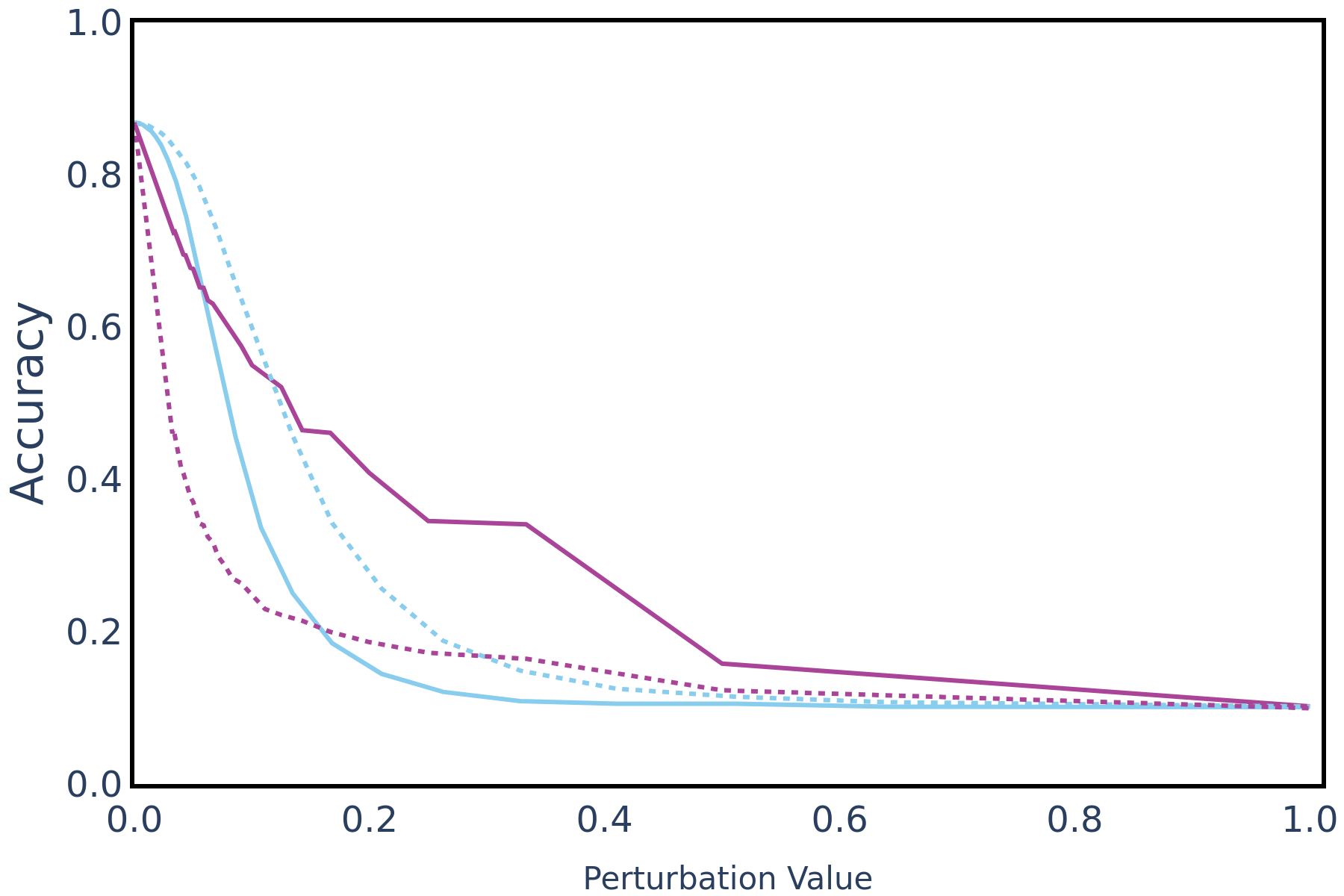}
    \end{subfigure}
    
    \begin{subfigure}[t]{0.49\textwidth}
        \caption{KMNIST\label{fig:subfigC3}}
        \includegraphics[width=\textwidth]{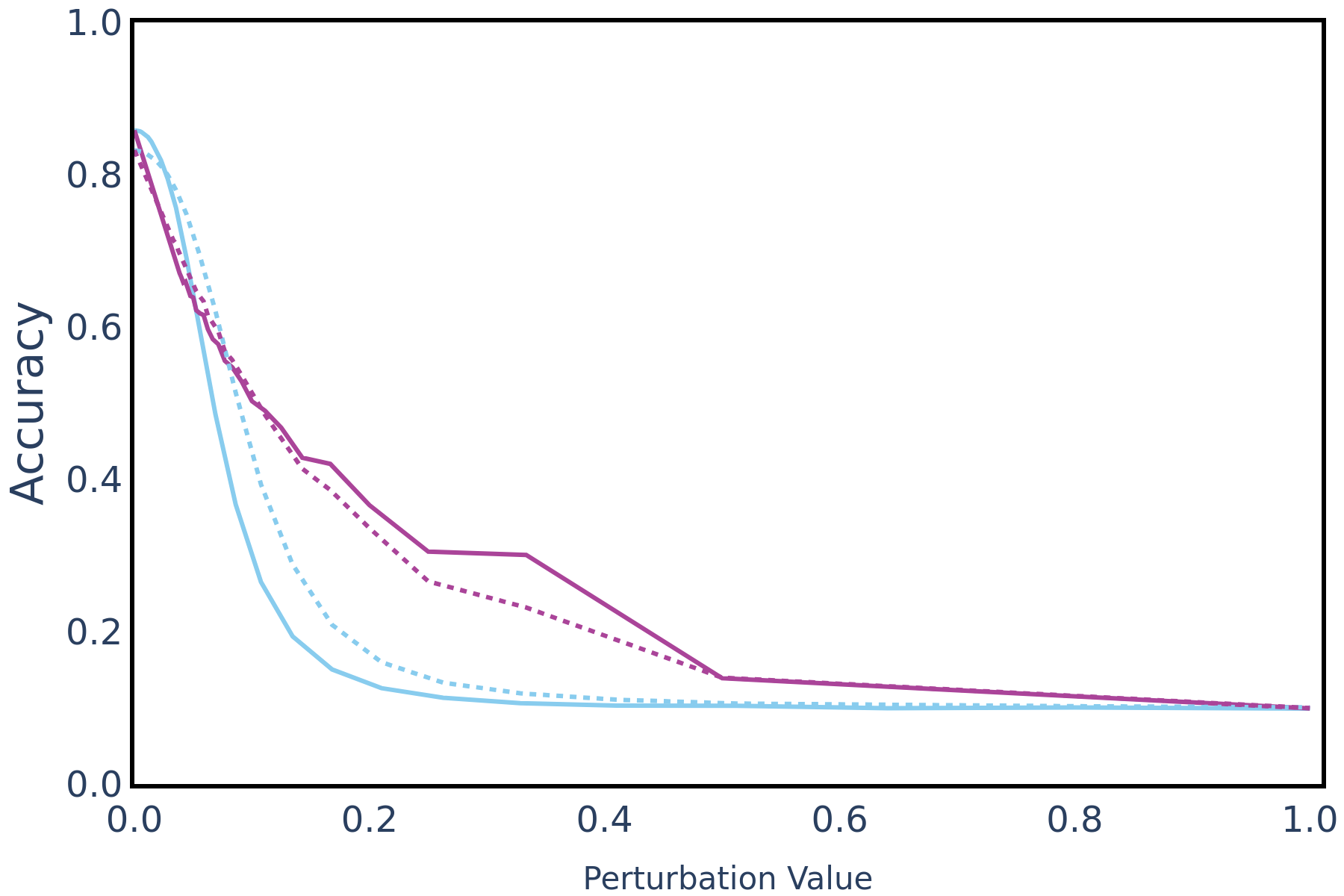}
    \end{subfigure}
    \begin{subfigure}[t]{0.49\textwidth}
        \caption{EMNIST (Letters)\label{fig:subfigD3}}
        \includegraphics[width=\textwidth]{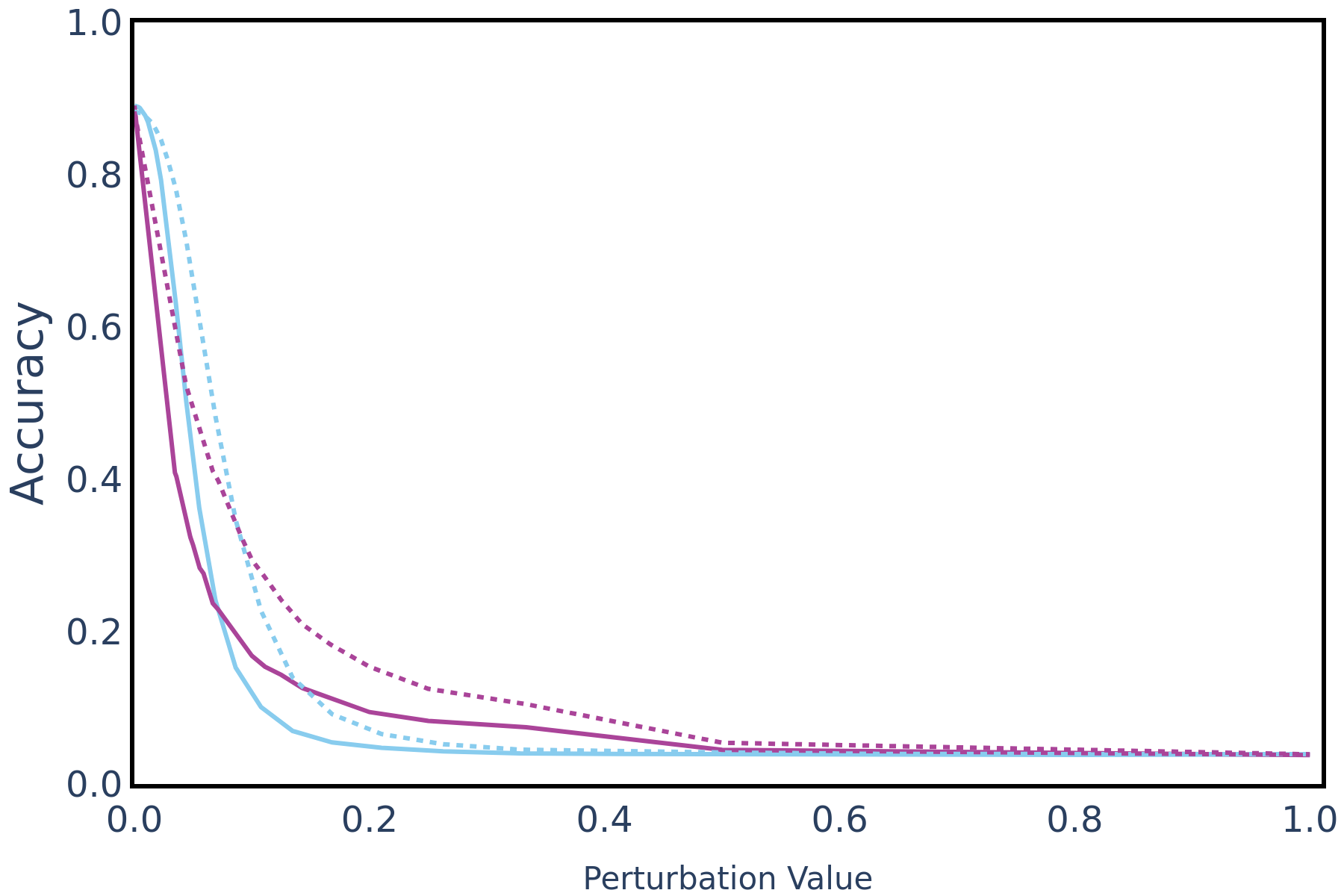}
    \end{subfigure}
    
    \begin{subfigure}[t]{\textwidth}
        \includegraphics[width=\textwidth]{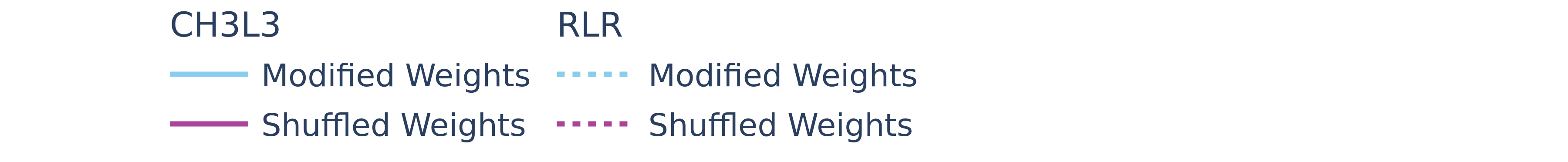}
    \end{subfigure}
    \caption{ \textbf{Robustness of trained sparse networks against perturbing the weights by shuffling or adding noise.} Accuracy as a function of the perturbation value for networks trained with the CH3L3 Link Regrowth method (continuous lines) and with the RLR method (dashed lines) for a) the MNIST dataset, b) the Fashion MNIST dataset, c) the KMNIST dataset, and d) the EMNIST letters dataset. Results for Weight Shuffling are shown in purple, for Weight Modification in light blue. }
    \label{fig:robustness_analysis_noising}
\end{figure}

In Fig.~\ref{fig:robustness_analysis_noising} we show the decay of accuracy as a function of the perturbation value for Weight Shuffling (purple) and Weight Modification (light blue). Similarly to Fig.~\ref{fig:robustness_analysis_pruning}, curves plotted with continuous lines correspond to the results obtained for the networks trained with the CH3L3 link regrowth method, whereas results shown with dashed lines were obtained with the RLR method. For all tested datasets, the accuracy drops very fast for both types of perturbation, with the Weight Shuffling reducing the accuracy slightly even faster than the Weight Modification at the very start of the perturbation process. However, at some point the curves cross and in the later stages the accuracy for Weight Shuffling remains higher compared to that of Weight Modification. According to Fig.\ref{fig:robustness_analysis_noising}, networks trained according to the CH3L3 method show slightly but noticeably stronger resilience against Weight Modification compared to networks trained with RLR in all tested datasets. In parallel, the comparison of the two training methods with respect to resilience against Weight Shuffling yields mixed results: Networks trained according to the CH3L3 method achieved higher robustness in the case of the MNIST and the FashionMNIST datasets (Fig.\ref{fig:robustness_analysis_noising}a-b), whereas in the case of the KMNIST and the EMNIST Letters datasets it is the other way around (Fig.\ref{fig:robustness_analysis_noising}c-d).

\captionsetup[subfigure]{justification=raggedright,singlelinecheck=off}

\captionsetup[subfigure]{justification=raggedright,singlelinecheck=off}
\begin{figure}[htbp]
    \centering
    \begin{subfigure}[t]{0.49\textwidth}
        \caption{MNIST\label{fig:subfigA4}}
        \includegraphics[width=\textwidth]{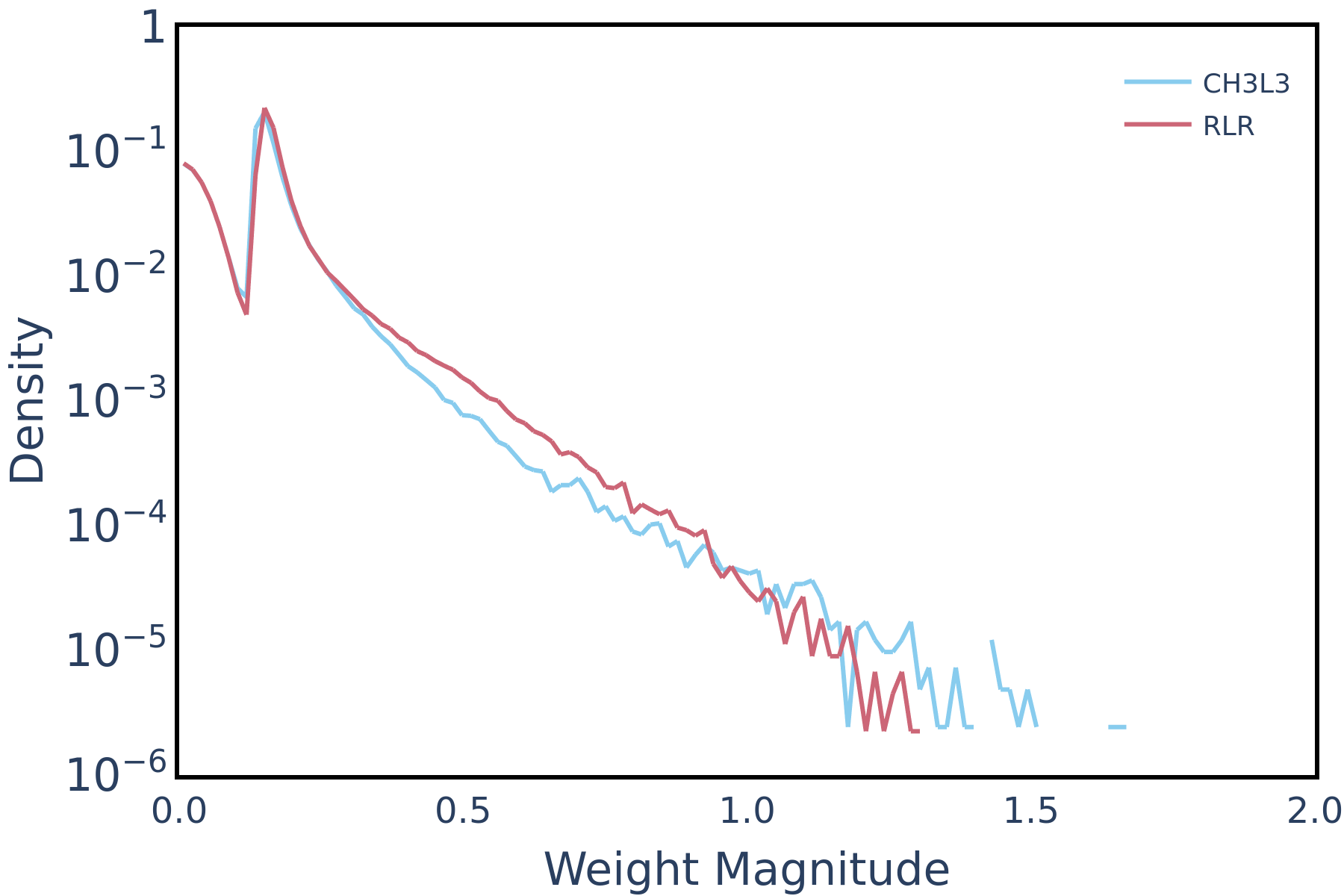}
    \end{subfigure}
    \begin{subfigure}[t]{0.49\textwidth}
        \caption{FashionMNIST\label{fig:subfigB4}}
        \includegraphics[width=\textwidth]{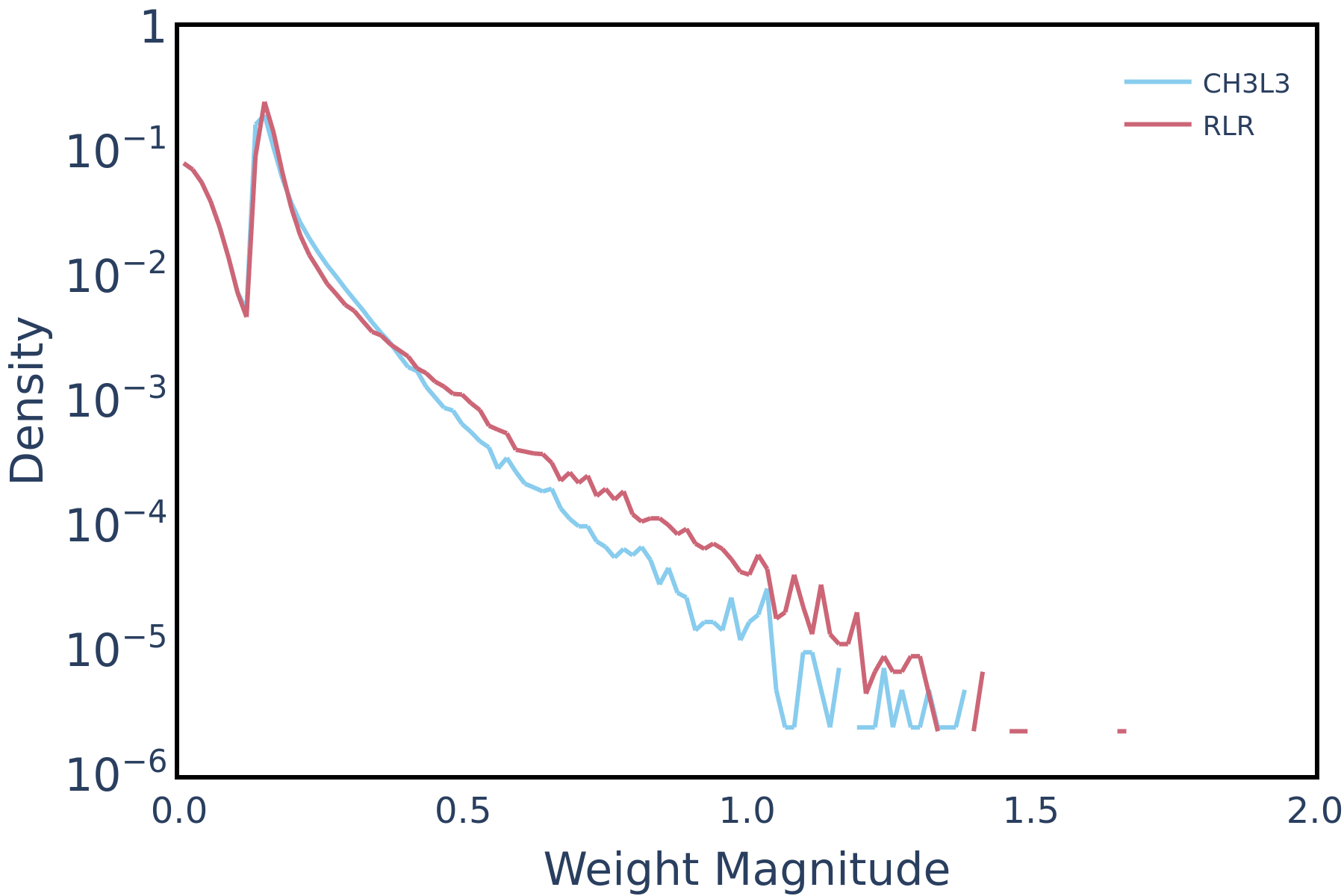}
    \end{subfigure}
    
    \begin{subfigure}[t]{0.49\textwidth}
        \caption{KMNIST\label{fig:subfigC4}}
        \includegraphics[width=\textwidth]{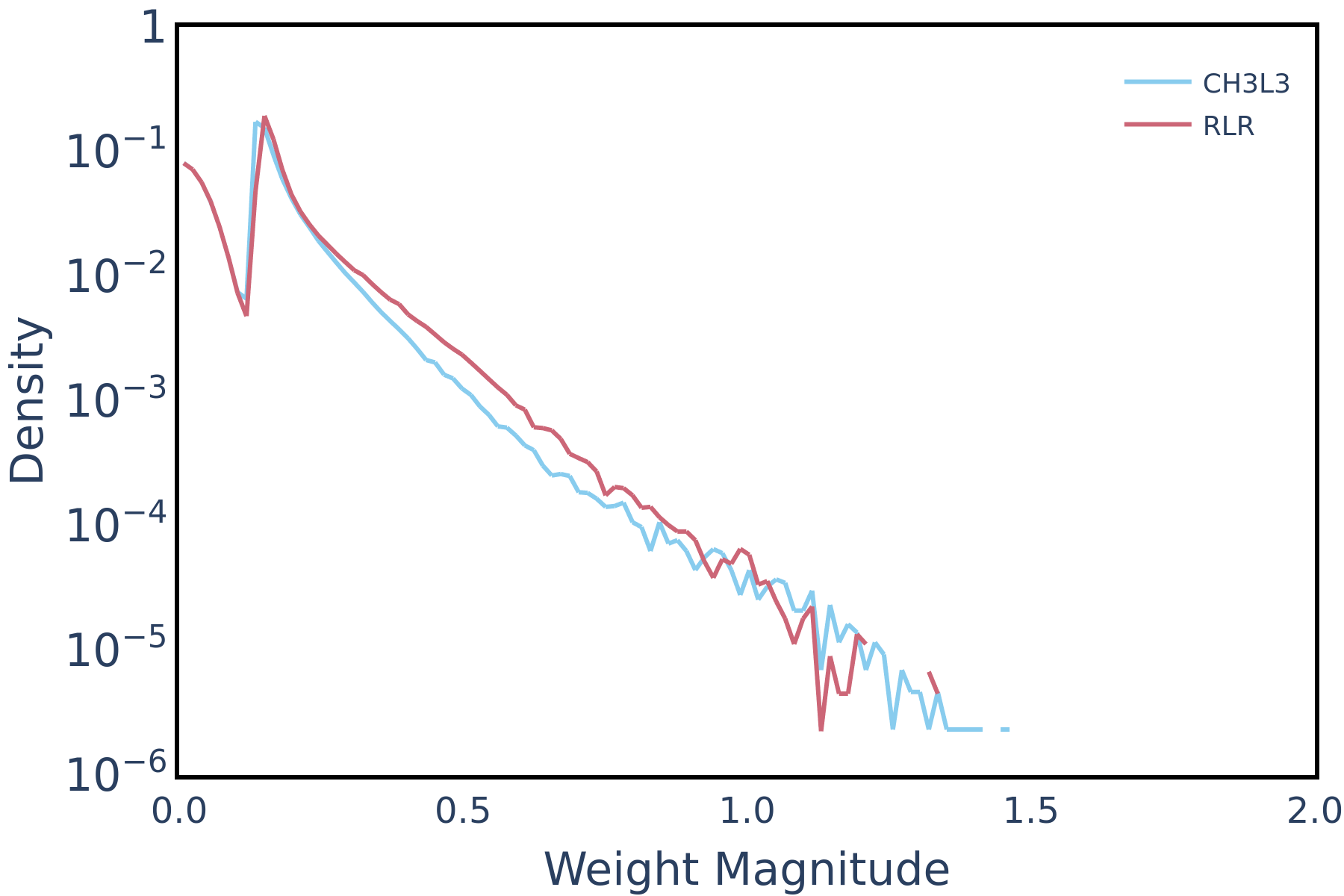}
    \end{subfigure}
    \begin{subfigure}[t]{0.49\textwidth}
        \caption{EMNIST (Letters)\label{fig:subfigD4}}
        \includegraphics[width=\textwidth]{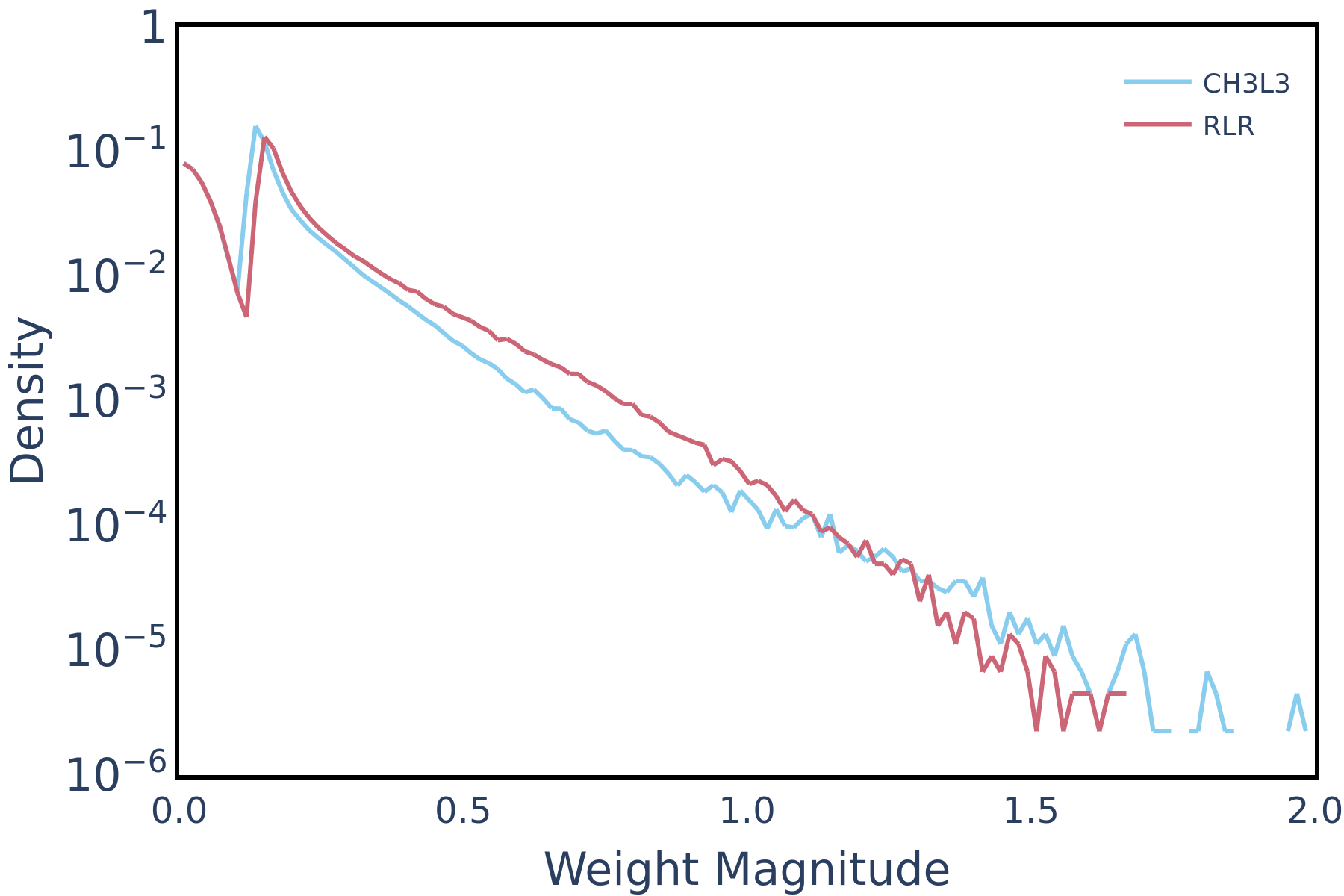}
    \end{subfigure}
    \caption{ {\bf Weight magnitude distribution in the trained networks.} We plot the distribution density for the absolute value of the connection weights at the end of the sparse training process, where results for networks trained according to the CH3L3 algorithm are shown in light blue and results for networks trained with RLR are shown in red. Panels depict the distributions observed for a) the MNIST dataset, b) the FashionMNIST dataset, c) the KMNIST dataset, and d) the EMNIST letters dataset.}
    \label{fig:total_weight_distribution}
\end{figure}

To investigate potential structural differences between networks trained using the CH3L3 algorithm and the RLR method, we measured the link weight distribution upon completion of the training processes, with Fig.~\ref{fig:total_weight_distribution} displaying the resulting density distributions of the weight magnitude across the various datasets.
For all datasets and both training methods, we observe a narrow and pronounced local minimum in the density distribution near the origin. This is a consequence of the repeated link removal in each epoch, where a fixed fraction of the connections --specifically the weakest links-- is consistently pruned. The steep decay immediately to the left of this minimum is primarily attributed to new links inserted in the final epoch, as these are initialized with weights drawn from a normal distribution centered at zero. Following this local minimum, a similarly narrow global maximum can be observed, and the density curves then proceed with a near-exponential decay towards larger weight values. When comparing the two training methods, the distributions show a consistent difference within the intermediate and moderately large weight range: specifically, in the range of approximately $w=0.2$ to $w=1$, the density curves for networks trained with RLR (red lines) consistently exceed those for networks trained with the CH3L3 algorithm (light blue lines) across all datasets. However, the results for extremely large and rare weights ($w > 1$) vary by dataset: the density for CH3L3 networks was higher than for RLR networks in the MNIST (Fig. 4a) and EMNIST datasets (Fig. 4d), was lower than for RLR networks in the Fashion MNIST dataset (Fig. 4b), and was roughly equal for the KMNIST dataset (Fig. 4c).

%The properties of the weight magnitude densities, taken together with the natural assumption that connections with large weights contribute more to the accuracy and performance of the neural network, provide a qualitative explanation for the some of the observed differences in the robustness of networks trained with different methods. Since a seemingly larger portion of the connections has medium to moderately large weight in networks trained with RLR compared to networks trained with the CH3L3 algorithm, when subject to Reverse Weight Order Pruning, at the same fraction of removed connections, the weights of the remaining connections are larger in a network trained with RLR, and thus, the network itself can be expected to perform better. However, when removing the links in a random fashion, it is the other way around, since we can expect a larger damage in terms of weights for RLR trained networks compared the CH3L3 trained networks at the same fraction of removed links.

The measured properties of the weight magnitude densities, coupled with the assumption that connections with large weights contribute more significantly to network accuracy, offer a qualitative explanation for the observed robustness differences. Networks trained with RLR exhibit a higher density of medium-to-moderately-large weights compared to CH3L3 networks. Consequently, when both network types are subjected to Reverse Weight Order Pruning, the RLR networks retain larger weights among their remaining connections for any given fraction of links removed. This structural difference is expected to translate into superior performance and resilience for RLR-trained networks under RWO Pruning. Conversely, when links are removed randomly, the higher density of important, medium-to-moderately-large weights in RLR networks makes them more vulnerable to damage, resulting in a larger performance degradation compared to the CH3L3-trained networks at the same removal fraction.

\section{Discussion}

The study of network robustness, often framed as an inverse percolation process, where the network structure is incrementally deteriorated through subsequent link removal, has a long history in network science\cite{Barabasi_attack_and_error_tolerance,Artime_Nat_Rev_Phys_2024}. Depending on the initial network structure and the specific perturbation applied, at a certain point the system may easily undergo a drastic change analogous to a phase transition in statistical physics. Such percolation transitions have been observed both in living neural networks, when chemically blocking neurotransmitter receptors~\cite{Ilan_PRL_97}, and in simulated neural networks, where resilience against various attacks strongly depended on the underlying architecture. The sequential pruning of connections is also a crucial method for compression in artificial neural networks, allowing an initially dense network to be converted into a sparse one while retaining near-original performance, despite the loss of the majority of its connections.

Relatedly, in the present paper we investigated the consequences of pruning and weight perturbations in neural networks that are already sparse from the outset. The networks studied were trained on four distinct datasets using two alternative dynamic training methods: the CH3L3 method (in which a fixed fraction of connections are removed and then reintroduced based on missing link prediction in each epoch) and the Random Link Regrowth (RLR) method (where the removed links are reintroduced at random).
%The studied networks were trained on 4 different datasets according to two alternative dynamic training methods: the CH3L3 method (where a fixed fraction of the connections are first removed and then reintroduced based on missing link prediction in each training epoch) and the Random Link Regrowth method (similar to the CH3L3, but the removed links are reintroduced at random). 

Analysis of the training dynamics revealed that, across all datasets, CH3L3 networks reached maximum accuracy after fewer training epochs compared to RLR networks, though the final convergence accuracy for both methods was consistently the same within a negligible margin.
Regarding resilience, CH3L3 networks proved to be more resilient against random link removal for three of the four datasets, with the fourth dataset showing nearly identical accuracy decay curves for both methods. In contrast, RLR networks showed higher resilience against Reverse Weight Order removal across all datasets. This latter result is particularly notable as it demonstrates the potential for significant further sparsification in RLR-trained networks, allowing up to approximately 80\% of the connections to be removed without a critical loss in accuracy. However, both network types were highly susceptible to Weight Order removal, with accuracy often reducing to its minimum value after the removal of only 3\% to 10\% of the links. Beyond link removal, we analysed the effects of direct weight modification, finding that RLR-trained networks exhibited slightly greater resilience against random noise added to the weight values, whereas CH3L3 networks proved more resilient against random weight shuffling in three out of the four tested datasets.

A qualitative explanation for these differences stems from the resulting weight distributions. Our results show that in RLR-trained networks, a higher proportion of the distribution is concentrated in the medium-to-high weight range compared to CH3L3 networks. Based on this structural difference, RLR networks are expected to show higher resilience under Reverse Weight Order Pruning (which preferentially removes small weights), while CH3L3 networks are expected to be more resilient against Random Pruning (which is more likely to disrupt the concentrated, critical weights in the RLR distribution).

In summary, our study of the robustness of dynamically trained sparse neural networks against link pruning and weight modification shows that the specific dynamic training method employed has a notable effect on network resilience. Furthermore, results reveal that RLR-trained networks possess a structural quality that allows for extensive post-training compressibility without a significant loss in accuracy.

%In summary, our studies of the robustness of dynamically trained sparse neural networks against link pruning and weight modification showed that the details of the training method can have a notable effect on the network resilience. Furthermore, the results also revealed that networks trained with RLR can be further sparsified to a large extent without a significant loss in the training accuracy.

\section{Methods}

Our approach is based on training sparse neural networks while dynamically modifying their connectivity structure during training. The overall procedure consists of two main components: (i) weight learning and (ii) topology update. These are applied in an alternating fashion throughout training.

\subsection{Architecture}
We employ multiple sparse layers followed by a single dense layer.
The size of the input layer is fixed by the problem we are solving, for instance the popular MNIST dataset consists of input images of $28 \times 28$ pixels, this corresponds to input size of $784$ after flattening the image.
This is followed by a certain number of hidden neuron layers, for all experiments in this study we used three layers of 1000 neurons unless otherwise specified. These hidden neurons are sparsely connected to the neurons of the preceding layer.
Finally, the last layer of hidden neurons is densely connected to the output layer whose size is again specified by the problem at hand. MNIST has 10 classes, therefore this last dense layer contains $1000 \times 10$ links.
Each neurons on the hidden layers represent an aggregation (weighted sum) of incoming links and a ReLU activation function \cite{nair2010rectified}.

\subsection{Initialization}
We begin by initializing a sparse neural network with a predefined sparsity level.
The initial connectivity pattern is created at random following the Erdős-Rényi random graph model.
Of course, any other method can be used here.
The weight parameters of the network are then initialized.
The Kaiming initialization \cite{he2015delving} is a popular option when working with ReLU activation, as it preserves the variance of activations during the forward pass and variance of gradients during the backward pass.
In practice this is achieved by sampling weight values from the normal distribution $N(0, 2/n_{in})$, where $n_{in}$ is the number of neurons of the previous layer (i.e. total number of possible incoming links).
We are using Kaiming initialization unless otherwise mentioned.

\subsection{Training Procedure}
\begin{figure}[htbp]
    \centering
    \begin{subfigure}[t]{0.32\textwidth}
        \caption{\label{fig:subfigA5}}
        \includegraphics[width=\textwidth]{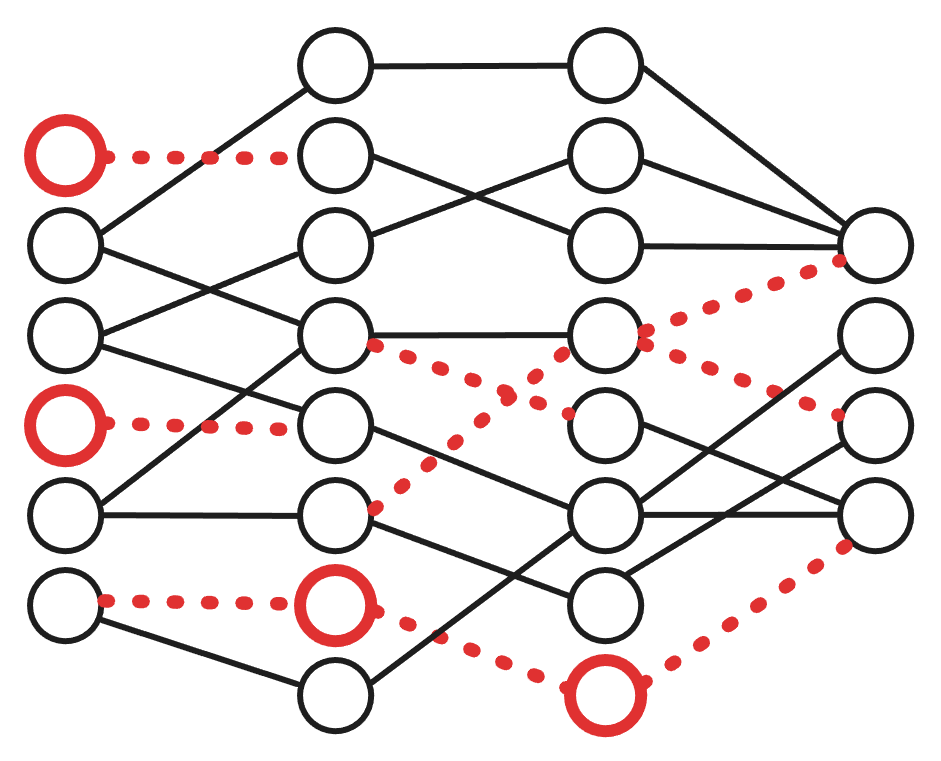}
    \end{subfigure}
    \begin{subfigure}[t]{0.32\textwidth}
        \caption{\label{fig:subfigB5}}
        \includegraphics[width=\textwidth]{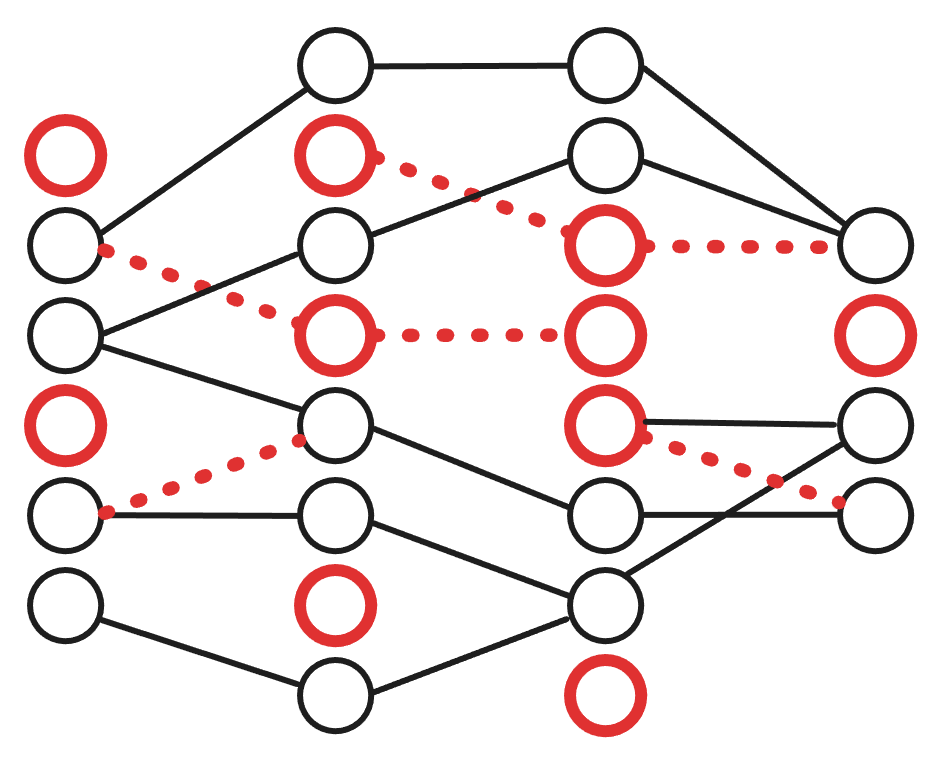}
    \end{subfigure}
    \begin{subfigure}[t]{0.32\textwidth}
        \caption{\label{fig:subfigC5}}
        \includegraphics[width=\textwidth]{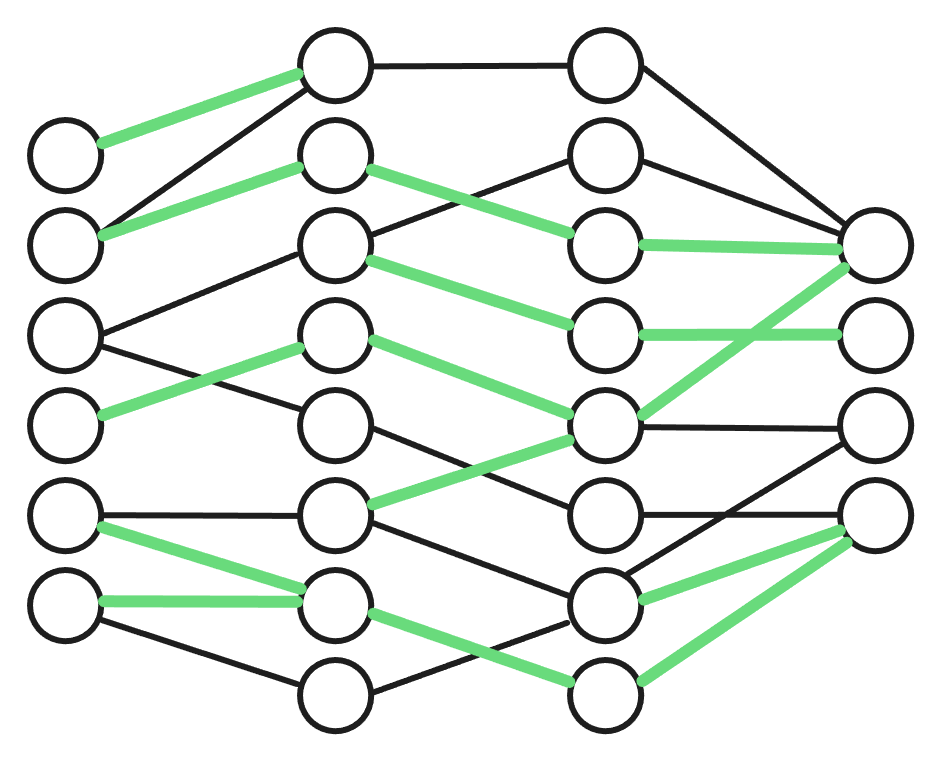}
    \end{subfigure}
    \caption{
   \textbf{The three stages of the topology update.}
    %\textbf{The topology update has three stages.} 
    In the first stage (a) a fraction of links (dotted red lines) is removed in each sparse layer. In the second optional stage (b) all links that are no longer connected to the input and/or output layer (dotted red lines) are also removed. In the final stage (c) new links (green) are drawn and initialized such that the total number of links in each layer will be unchanged after the procedure.}
    \label{fig:topology_update}
\end{figure}

Training takes place in discrete epochs. Each epoch is composed of the following steps:

\begin{enumerate}
    \item \textbf{Weight learning.} Given the current sparse connectivity, standard gradient-based optimization is applied to update the active parameters.
    During this step the network adapts its weight values according to the task at hand, conditioned on the present topology.
    
    \item \textbf{Topology update.} After each epoch, the connectivity of the network is modified according to a specified update rule.
    The stages of the topology update are shown in Fig.~\ref{fig:topology_update}.
    
    One always starts with pruning (removing) links that are deemed irrelevant.
    In the simplest case we remove a predefined fraction of all existing links based on the magnitude of their weight values.
    This is motivated by the fact that connections with vanishing weight values tend to not contribute meaningfully to the activation of the neuron they feed into.
    
    During an optional second pruning stage, we remove all disconnected neurons and their associated links.
    A neuron is considered disconnected if it does not lie on any directed path originating from the input layer and ending on the output layer.
    A link that is not connected to the output simply occupies precious space in the sparse network that could be used to house a meaningful link.
    Similarly, a link not connected to the input will at best have contributed to the activation of the neuron by a constant offset and should therefore be merged with the appropriate bias neuron.

    Then, new links are grown.
    One may opt to introduce new links randomly or according to some heuristic.
    We analyzed trained sparse networks with a random rule -- this is the method used in [PAPER FOR S.E.T. METHOD] -- CH3L3 link prediction [PAPER for CH3L3] as well as simpler link prediction method based on counting the number of paths connecting two neurons.

    Finally, the new links are initialized by sampling from a distribution.
    Here, again, we use Kaiming initialization unless otherwise specified.
\end{enumerate}

By alternating between weight learning and topology updates, the network explores different sparse configurations over the course of training, potentially improving its ability to discover efficient and expressive structures.

\captionsetup[subfigure]{justification=raggedright,singlelinecheck=off}

\bibliographystyle{unsrt}
\bibliography{Refs}

\end{document}